\RequirePackage{amsmath}
\documentclass[runningheads]{llncs}
\usepackage[T1]{fontenc}
\usepackage{amsfonts}  
\usepackage{algorithm}
\usepackage{algorithmic}
\usepackage{booktabs}
\usepackage{multirow}
\usepackage{graphicx}
\usepackage{xcolor}
\usepackage{hyperref}
\hypersetup{
  hidelinks,
}
\usepackage[caption=false]{subfig}
\usepackage{appendix}

\newcommand{\frameworkName}{Concept Discovery through Latent Diffusion-based Counterfactual Trajectories}
\newcommand{\frameworkAcronym}{CDCT}

\begin{document}

\title{Generating Counterfactual Trajectories with Latent Diffusion Models for Concept Discovery}

\titlerunning{Counterfactual Trajectories for Concept Discovery}

\author{Payal Varshney\textsuperscript{*,}\inst{1,2}\orcidID{0009-0009-6571-6134} \and
Adriano Lucieri\textsuperscript{*,}\inst{1,2}\orcidID{0000-0003-1473-4745} \and
Christoph Balada\inst{1,2}\orcidID{0000-0003-0307-7866} \and
Andreas Dengel\inst{1,2}\orcidID{0000-0002-6100-8255} \and
Sheraz Ahmed\inst{2}\orcidID{0000-0002-4239-6520}
}

\def\thefootnote{*}\NoHyper\footnotetext{These authors contributed equally to this work.}\endNoHyper\def\thefootnote{\arabic{footnote}}
\authorrunning{P. Varshney et al.}

\institute{Rheinland-Pfälzische Technische Universität Kaiserslautern-Landau,\\Kaiserslautern, Germany \and
German Research Center for Artificial Intelligence GmbH (DFKI),\\Kaiserslautern, Germany\\
\email{firstname.lastname@dfki.de}\\
}
\maketitle
\begin{abstract}
Trustworthiness is a major prerequisite for the safe application of opaque deep learning models in high-stakes domains like medicine.
Understanding the decision-making process not only contributes to fostering trust but might also reveal previously unknown decision criteria of complex models that could advance the state of medical research.
The discovery of decision-relevant concepts from black box models is a particularly challenging task.
This study proposes \textit{\frameworkName} (\frameworkAcronym), a novel three-step framework for concept discovery leveraging the superior image synthesis capabilities of diffusion models.
In the first step, \frameworkAcronym\ uses a Latent Diffusion Model (LDM) to generate a counterfactual trajectory dataset.
This dataset is used to derive a disentangled representation of classification-relevant concepts using a Variational Autoencoder (VAE).
Finally, a search algorithm is applied to identify relevant concepts in the disentangled latent space.
The application of \frameworkAcronym\ to a classifier trained on the largest public skin lesion dataset revealed not only the presence of several biases but 
also meaningful biomarkers.
Moreover, the counterfactuals generated within \frameworkAcronym\ show better FID scores than those produced by a previously established state-of-the-art method, while being 12 times more resource-efficient.
Unsupervised concept discovery holds great potential for the application of trustworthy AI and the further development of human knowledge in various domains.
\frameworkAcronym\ represents a further step in this direction.
\keywords{Explainablility \and Counterfactuals\and Concept Based Explanations \and Latent Diffusion Models \and Dermoscopy \and Concept Discovery}
\end{abstract}

\section{Introduction} \label{sec:introduction}

Deep learning (DL) algorithms have gained immense popularity for their outstanding performance in the past decade~\cite{rank2020deep,sturman2020deep,ranjan2018deep}.
The inherent non-linearity and over-parametrization of these algorithms make it particularly challenging to comprehend their reasoning processes.
This, among other factors, contributes to a general lack of trust in such black-box models, constituting a major impediment to their application in safety-critical domains such as medicine.
With the General Data Protection Regulation~\cite{voigt2017eu} and the upcoming European Artificial Intelligence (AI) Act~\cite{veale2021demystifying}, DL-based AI systems must now also comply with regulatory requirements directly concerning the topic of trust.
Research on eXplainable AI (XAI) aims at increasing trust by providing more profound insights into the decision-making of opaque black-box models.

The correct interpretation of an explanation by the explainee is fundamental for yielding an understanding of the opaque decision-making process, which is a crucial prerequisite for building trust~\cite{palacio2021xai}. 
Many popular XAI methods only provide information about the relevance of individual input features~\cite{ribeiro2016should,zhou2016learning,selvaraju2017grad}.
The interpretation of these methods by explainees often suffers from a lack of context about the informative value of feature relevance. 
Concept-based explanation methods promise to remedy this by allowing to localize~\cite{lucieri2020explaining} human-aligned concepts and quantify~\cite{kim2018interpretability} their relevance for decision-making.
However, the acquisition of fine-grained concept annotations is expensive and inflexible.
First efforts have been made towards unsupervised concept discovery~\cite{ghorbani2019towards,lang2021explaining,atad2022chexplaining}, which could reduce the burden of collecting labels for concepts but also comes with the prospect of discovering new, previously unknown concepts.
Existing methods are either based on strong structural assumptions~\cite{ghorbani2019towards} or make use of 
Generative Adversarial Networks (GANs)~\cite{lang2021explaining,atad2022chexplaining} that are usually challenging to train.

This work proposes \textit{\frameworkName} (\frameworkAcronym), the first concept discovery framework based on counterfactual trajectories generated by state-of-the-art latent diffusion models (LDMs)~\cite{rombach2022high}.
The first step encompasses the generation of a counterfactual trajectory dataset using a text-conditioned LDM guided by the target classifier. 
This step aims to extract relevant semantic changes that describe the classifier's decision boundaries.
The counterfactual trajectory dataset is used in the subsequent step to compute a disentangled representation of the classifier-relevant features using a Variational Autoencoder (VAE).
In the final step, the VAE's latent space is exploited for concept discovery through a search algorithm that is built upon~\cite{lang2021explaining}.
The contribution of our work is as follows:
\begin{itemize}
     \item we propose \frameworkAcronym, the first concept discovery framework that leverages latent diffusion-based counterfactual trajectories.
     \item we demonstrate improved counterfactual explanation capabilities by combining latent diffusion models with classifier guidance.
     \item we apply \frameworkAcronym\ to a skin lesion classifier, demonstrating its ability to reveal classifier-specific concepts including biases and new biomarkers.
\end{itemize}

\section{Related Work} \label{sec:relatedwork}
The field of eXplainable AI can be subdivided by the output modality of a given XAI method.
Several popular examples in the image domain are based on the quantification of feature relevance through surrogate models~\cite{ribeiro2016should}, activations~\cite{zhou2016learning}, or gradients~\cite{selvaraju2017grad}.
Recently, human-centric XAI methods like counterfactuals~\cite{wachter2017counterfactual,dandl2020multi,van2021interpretable} or concept-based explanation methods~\cite{kim2018interpretability,koh2020concept} struck particular interest in the community, as they focus primarily on facilitating the interpretation of explanations by stakeholders and thus achieving their trust.

\subsection{Generative Counterfactual Explanations}
Counterfactuals constitute alternative classification scenarios with different outcomes given a marginal modification of a reference image.
Wachter et al.~\cite{wachter2017counterfactual} first introduced a method for the generation of counterfactuals through optimization.

The closest related works use diffusion models (DMs) to generate counterfactuals~\cite{jeanneret2022diffusion,sanchez2022diffusion,augustin2022diffusion,sanchez2022healthy,jeanneret2023adversarial}.
Some works use an unconditional diffusion model for the generation of counterfactuals guided by the target classifier~\cite{jeanneret2022diffusion,sanchez2022diffusion}.
Augustin et al.~\cite{augustin2022diffusion} utilize adaptive parameterization and cone regularization of gradients.
Sanchez et al.~\cite{sanchez2022healthy} made use of a conditional diffusion model to generate healthy counterfactual examples of brain images to localize abnormal lesions.
In~\cite{jeanneret2023adversarial}, an adversarial pre-explanation is improved by diffusion-based inpainting to generate minimal counterfactuals.
All previous works on DM-based counterfactual generation operate in the pixel space.
Our proposed work instead leverages the more recent latent diffusion model to make use of improved synthesis capabilities with better computational efficiency, while focusing on semantic changes through guidance in the latent instead of the pixel space.  

\subsection{Concept-based Explainability}
Concept-based XAI methods aim at the quantification and localization of higher-level concepts that are aligned with human cognitive processes. 
Concept-based explanations in pedestrian detection might for instance highlight a group of pixels as belonging to the concept \textit{face}.
Supervised concept-based explanation methods~\cite{kim2018interpretability,lucieri2020explaining,koh2020concept,espinosa2022concept} rely on concept annotations, which can be expensive to obtain. Moreover, in complex domains like medicine, often experts are required for such annotations.
However, some works approached the unsupervised discovery of novel concepts.
One line of work deals with the identification of concepts from a classifier's latent representation~\cite{ghorbani2019towards,wang2023learning,zhang2021invertible,fel2023craft,vielhaben2023multi}.
Ghorbani et al.~\cite{ghorbani2019towards}, for instance, proposed a framework for concept discovery based on the segmentation of images, clustering of their latent representations, and quantification of the conceptual influence of clusters.
Fel et al.~\cite{fel2023craft} extend the work of~\cite{ghorbani2019towards} and~\cite{zhang2021invertible} to recursively decompose concepts across layers and localize them.
In another approach, Achtibat et al.~\cite{achtibat2023attribution} introduced Concept Relevance Propagation (CRP), combining local and global perspectives to address both the ‘where’ and ‘what’ questions for individual predictions.
Additionally, Poeta et al.~\cite{poeta2023concept} provided a comprehensive survey on concept-based explainable AI, detailing various methodologies and their applications.

Another relevant line of work regards the classifier as a black box and aims at the discovery of concepts through the generative manipulation of input samples~\cite{lang2021explaining,atad2022chexplaining,ghandeharioun2021dissect,song2023img2tab}.
Lang et al.~\cite{lang2021explaining}, for instance, modified the StyleGAN architecture to systematically examine the GAN's style space for class-influential concept dimensions.
Ghandeharioun et al.~\cite{ghandeharioun2021dissect} approach concept discovery by enforcing the disentanglement of a GAN's latent concepts, encouraging higher distances between dissimilar and proximity between similar concepts.
In contrast to existing concept discovery frameworks, \frameworkAcronym\ leverages the power of state-of-the-art, conditional latent diffusion models for the generation of counterfactual trajectories to disentangle and identify classifier-relevant information.


\section{Methodology} \label{sec:methodology}
\begin{figure}[h!]
    \centering
    \includegraphics[width=1\linewidth]{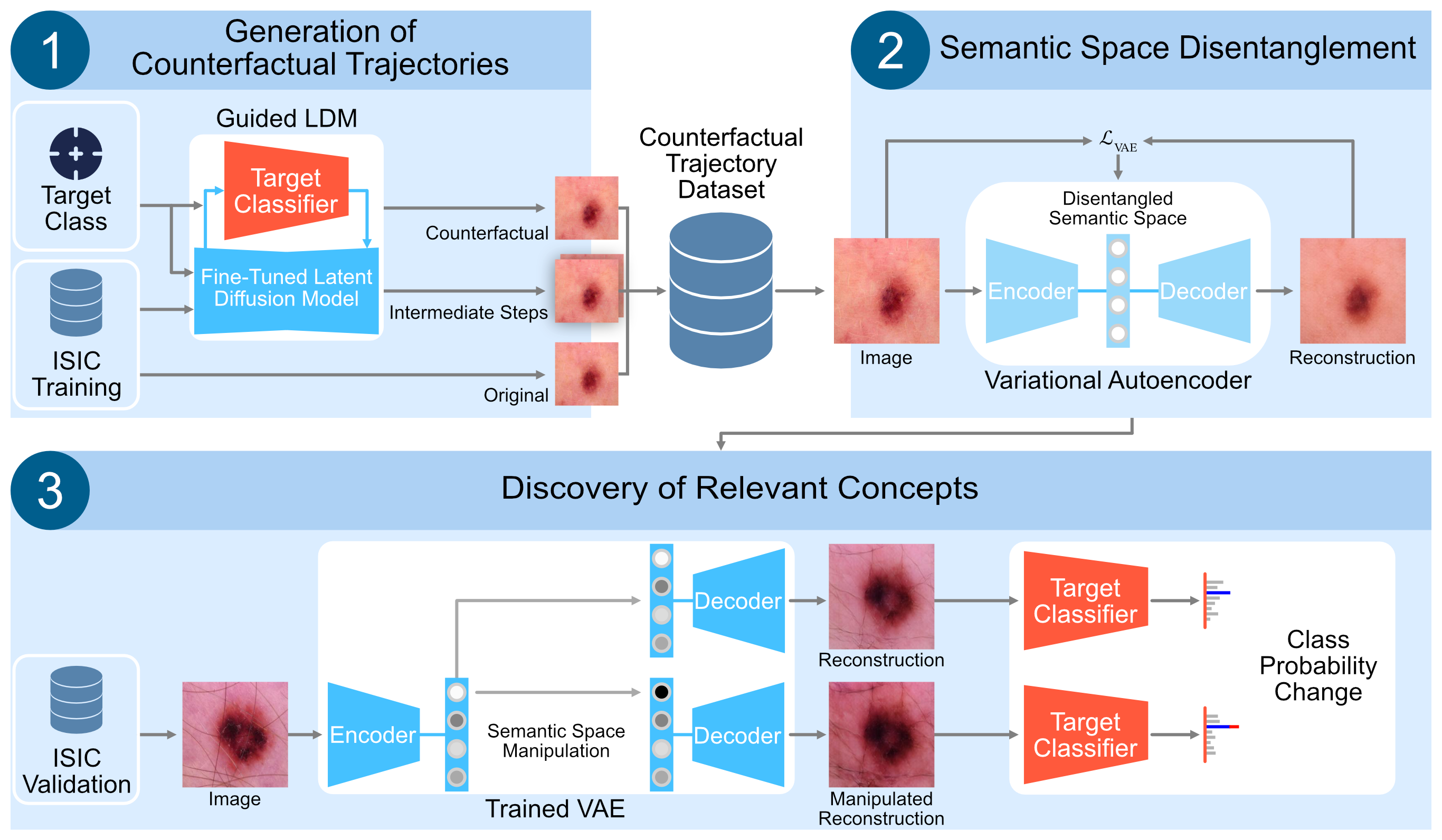}
    \caption{\frameworkAcronym\ is a three-step concept discovery framework. An LDM with classifier guidance is used to generate a counterfactual trajectory dataset. A VAE is trained on this trajectory dataset to disentangle decision-relevant features. Finally, class-relevant dimensions are identified by manipulating the VAE's latent space and observing the target classifier's output.}
    \label{fig:Concept_discover_framework}
\end{figure}

\frameworkAcronym\ is a three-step framework for unsupervised concept discovery. 
In the first step, a latent diffusion model with classifier guidance is used to generate a counterfactual trajectory dataset, capturing decision-relevant concepts of the target classifier.
This dataset is used to obtain a disentangled representation of concepts with the help of a Variational Autoencoder. 
Finally, decision-relevant dimensions are identified within the VAE by analyzing their impact on the classifier's output. 
An overview of the framework is presented in figure~\ref{fig:Concept_discover_framework}.

\subsection{Generation of Counterfactual Trajectories}
\label{sec:methodology:step1}
\begin{figure}[h]
\centering
    \includegraphics[width=\textwidth]{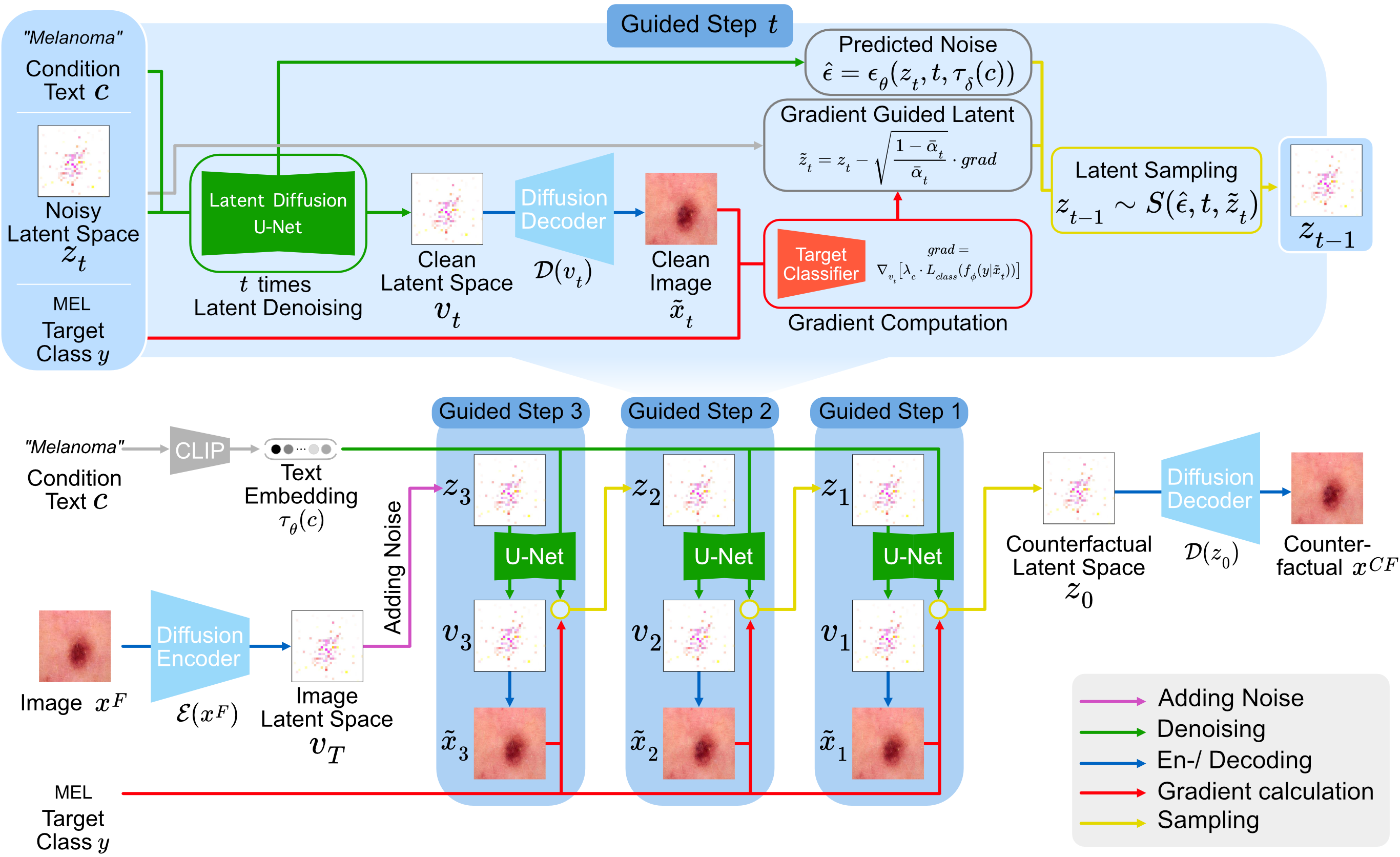}
        \caption{The counterfactual generation process starts by encoding the image $\mathcal{E}(x^F)$ and perturbing it to obtain $z_T$ (here $T=3$). The text encoder transforms the condition $(c)$ into an embedding $\tau_\theta(c)$. 
    Both $z_T$ and $\tau_\theta(c)$ are fed to the diffusion model for denoising.
    At guided step $t$, we denoise the noisy latent $z_t$, $t$ times to produce the clean latent $v_t$, which is decoded into a clean image $\tilde{x_t}$ to calculate the gradient of loss $\mathcal{L_{\text{class}}}$ for updating $z_t$. 
    We sample the previous less noisy latent $z_{t-1}$ from the estimated noise $\hat\epsilon$ and the updated noisy latent $\tilde{z_t}$. 
    In the final guided step, $z_0$ is decoded to yield the counterfactual image $(x^{CF})$.}
    \label{fig:counterfactual}
\label{fig:part1}
\end{figure}

The first step aims to extract decision-relevant variations of the input image from the target classifier into a counterfactual trajectory dataset. 
In contrast to previous works~\cite{jeanneret2022diffusion,augustin2022diffusion,sanchez2022healthy}, our proposed framework utilizes a latent diffusion model for a counterfactual generation to benefit from the improved generation efficiency and quality. 
Text conditioning is used to structure the latent space in accordance with known concepts.
Figure \ref{fig:counterfactual} depicts the counterfactual generation process with detailed information about the guided step $t$.

We start the process with an original image $x^F$ and utilize the encoder $\mathcal{E}$ to transform it into a lower-dimensional latent space. In the forward process, we take its latent representation to calculate a noisy version $z_t$ with variance schedule $\beta_t \in (0,1)$, where $\alpha_t = 1 - \beta_t$, $\bar\alpha_t = \prod_{k=1}^{t}\alpha_k$ and $ 1\; \leq \;  t\;  \leq \; T$. 

\begin{equation}
\label{eq:noisy_image}
    z_T = \sqrt{\bar\alpha_t} \; \mathcal{E}(x^{F}) + \epsilon_t \; \sqrt{1-\bar\alpha_t},\quad\textrm{where}\quad\epsilon_t\sim\mathcal{N}(\mathbf{0},\mathbf{I})
\end{equation}

Subsequently, we traverse the reverse Markov chain to generate a counterfactual with adapted classifier guidance introduced by Jeanneret et al.~\cite{jeanneret2022diffusion} applied to the latent space. In the reverse process, we iteratively go through the following steps until $t = 0$: 
\begin{enumerate}
  \item The LDM produces the cleaned latent vector $v_t$ by denoising $z_t$, $t$ times. 
  \item Then, $v_t$ is decoded to obtain its representation in the image space $\tilde{x}_t$.
  \item The target classifier $f_\phi$ is then applied on $\tilde{x}_t$ to calculate the gradients of the classification loss $\mathcal{L_{\text{class}}}$.
    \begin{equation}
    \label {eq:clean-grads} 
    \nabla_{z_t}L(z_t, y) = \frac {1}{\sqrt {\bar\alpha_t}} \nabla_{{v}_t}\big[\lambda_{c} \cdot L_{class}(f_\phi(y|\tilde{x}_t)) \big]
\end{equation}
  \item The noisy latent $z_t$ is guided using the gradient to produce $\tilde{z_t}$.
  \item $z_{t-1}$ is finally computed from $\tilde{z_t}$ and the noise $\hat\epsilon$ predicted at timestamp $t$.
\end{enumerate}

We utilize the PNDM~\cite{liu2022pseudo} sampling method represented as \(S(\hat{\epsilon}, t, \tilde{{z}_t}) \rightarrow z_{t-1}\) to get the less noisy sample at each timestamp $t$.
After $t$ repetitions of the aforementioned steps, the counterfactual image is generated by decoding $z_0$ using the decoder $x^{CF} = \mathcal{D}(z_0)$.
A sequence of clean images produced throughout the guiding process, commencing with the factual image and concluding with the counterfactual image, is referred to as a counterfactual trajectory.

\subsection{Semantic Space Disentanglement}

Counterfactual explanations may simultaneously alter multiple features, complicating the discovery of single, class-relevant concepts. 
Therefore, the counterfactual trajectory dataset is used in the second step to derive a disentangled representation of decision-relevant biomarkers for the target classifier.

Variational Autoencoders~\cite{kingma2013auto} are known for their ability to learn disentangled representations~\cite{chen2018isolating}.
To enhance the reconstruction fidelity of the VAE, three additional loss functions are integrated besides the default reconstruction loss ($\mathcal{L}_{\text{Rec}}$) and the regularization term ($\mathcal{L}_{\text{KLD}}$).
The L1 loss ($\mathcal{L}_{\text{L1}}$) is added to minimize pixel-wise differences, while the Structural Similarity Index Measure (SSIM)~\cite{zhao2015loss} ($\mathcal{L}_{\text{SSIM}} $) ensures perceptual image similarity.
Lastly, a perceptual loss ($\mathcal{L}_{\text{Perc}}$) based on all the layers of a pre-trained VGG19 is used to capture high-level semantic features.
Weighting the KLD loss by $w_{kld}$ balances the trade-off between reconstruction fidelity and disentanglement.
The final objective function for training the VAE can be stated as:
\begin{equation}
    \mathcal{L}_{\text{VAE}} = \mathcal{L}_{\text{Rec}} + w_{kld} \cdot \mathcal{L}_{\text{KLD}} +\mathcal{L}_{\text{L1}} + \mathcal{L}_{\text{SSIM}} + \mathcal{L}_{\text{Perc}} \nonumber
\end{equation}
The VAE is trained on the counterfactual trajectory dataset to disentangle the decision-relevant semantic features.

\subsection{Discovery of Relevant Concepts}
\label{sec:methodology:step3}
Finally, our framework modifies the AttFind~\cite{lang2021explaining} algorithm to identify decision-relevant concepts in the disentangled latent representation.
Algorithm~\ref{alg:part3} identifies the top latent space dimensions $L_y$ for a class with a given set of images. These dimensions yield the highest average increase in class probability when changing its value during reconstruction through the VAE.
In contrast to~\cite{lang2021explaining}, our approach modifies directions in the range $D_y \in [-3, +3]$ to account for the standard normal distribution, as high magnitudes do not yield artifacts or unrealistic changes but lead to properly emphasized concepts.
Moreover, concepts that yield inconsistent change directions were not filtered out, as we argue that conceptual manifestation is not necessarily monotonous on a linear trajectory.

\begin{algorithm}[h]
    \begin{algorithmic}    
        \STATE \textbf{Input:} Target classifier $f_\phi$, encoder $\mathcal{E}$, decoder $\mathcal{D}$,\\ Set $X$ of images with  $\forall x \in X, \quad f_\phi(x) \neq y$,         
        \STATE \textbf{Output:} Set $L_y$ of top latent space dimensions \& set $D_y$ of their directions.         
        \FOR{$x$ in $X$}        
        	\STATE $l \leftarrow \mathcal{E}(x), \quad l \in \mathbb{R}^m\;$ \quad // \text{Encoding}        
            \FOR{$i=1,\; \dots,\; l$}                      
                \FOR{$d$ in $[-3,\; -2, \;-1,\; 0, \;1,\; 2, \;3]$}                
                    \STATE $l[i] \;=\; d$        
                    \STATE $\tilde{x} \leftarrow \mathcal{D}(l)$ \quad // \text{Decoding}                     
                    \STATE Set $\Delta[x, l, d] = f_{\phi_y}(\tilde{x}) - f_{\phi_y}(\mathcal{D}(\mathcal{E}(x)))$ \quad //\text{Diff. softmax probability class $y$}                    
                \ENDFOR                
            \ENDFOR            
        \ENDFOR        
        \STATE Set $\bar\Delta[l, d] = \mathrm{Mean}(\Delta[x, l, d])$ over all $x \in X$        
        \STATE $L_y, D_y \leftarrow \text{argsort}(\bar\Delta[l, d], \text{descending order, top-K})$
    \end{algorithmic}
    \caption{Identify Relevant Latent Space Dimensions for a Specific Class}\label{alg:part3}
\end{algorithm}

\section{Experiments \& Results} \label{sec:experiments}

\subsection{Datasets \& Classification Models}\label{ssec:dataset}

Data from the International Skin Imaging Collaboration (ISIC) challenges (2016-2020)\footnote{The data is available at \url{https://challenge.isic-archive.com/challenges/}.} is used for experimentation.
The duplicate removal strategy proposed in Cassidy et al.~\cite{cassidy2022analysis} is followed, resulting in a consolidated dataset of 29,468 samples, further split into train, validation, and test sets.
The dataset consists of eight distinct classes, namely \textit{Melanocytic Nevus} (NV), \textit{Squamous Cell Carcinoma} (SCC), \textit{Benign Keratosis} (BKL), \textit{Actinic Keratosis} (AK), \textit{Basal Cell Carcinoma} (BCC), \textit{Melanoma} (MEL), \textit{Dermatofibroma} (DF), and \textit{Vascular Lesions} (VASC).
The ISIC 2016, ISIC 2017, PH2~\cite{mendoncca2013ph}, and Derm7pt~\cite{kawahara2018seven} datasets were used for the training of concept classifiers to provide missing conditioning labels.
The selected dermoscopic concepts are \textit{Streaks}, \textit{Pigment Networks}, \textit{Dots \& Globules}, \textit{Blue-Whitish Veils}, \textit{Regression Structures}, and \textit{Vascular Structures}.

In this study, a ResNet50~\cite{he2016deep} is trained on the consolidated ISIC training dataset, serving as the target classifier.
The individual concept classifiers are based on the target classifier, fine-tuned on the respective concept dataset. 

\subsection{Generation of Counterfactual Trajectories}
Stable Diffusion (SD) 2.1\footnote{Available at \url{https://huggingface.co/stabilityai/stable-diffusion-2-1}.} is used as a baseline for the latent diffusion model.
To enhance conditioning information for fine-tuning, captions are generated for training images including diagnostic details and skin lesion concepts. 
Concept classifiers were used to predict the presence or absence of concepts wherever no annotations were available.
Only present concepts were used as conditioning information.
The SD model is fine-tuned on the consolidated ISIC training dataset with text conditioning for 15K steps and a learning rate of $1e^{-5}$, generating realistic skin lesion samples.

\begin{figure}[!h]
    \centering
    \includegraphics[width=1\linewidth]{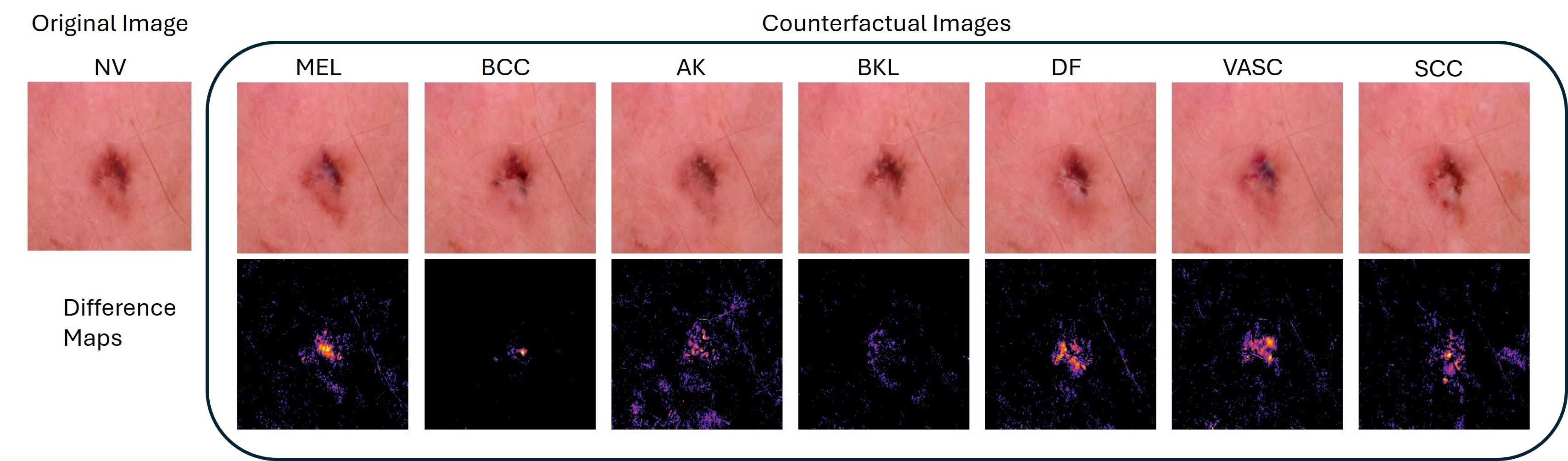}
    \caption{An image of the \textit{Nevus} class alongside its counterfactual images in all other target classes. The second row shows difference maps, providing an easy way to identify the areas of alteration between the original image and each counterfactual.}
    \label{fig:CF_all_cls}
\end{figure}

The methodology described in section~\ref{sec:methodology:step1} is applied to generate counterfactuals.
Experimentation revealed that a gradient loss scale of $\lambda_c > 10$ introduces unrealistic artifacts on the counterfactual images.
A value of $\lambda_c = 4$ has therefore been chosen for all results.
High values for the initial noise level $t$ compromised the image quality, while $t = 10$ yielded a good trade-off between generative capacity and image quality. 
Figure \ref{fig:CF_all_cls} shows exemplary counterfactuals from one source image to all other target classes, showcasing the ability of the LDM to generate class-selective biomarkers. 
More examples can be found in the supplementary material~\ref{sup:counterfactuals}.

Counterfactuals generated by different versions of \frameworkAcronym\ are quantitatively compared with the results from DiME~\cite{jeanneret2023adversarial} in table~\ref{tab:fid}.
The flip ratio measures the frequency at which generated counterfactual images are classified as the target class by the classification model.
A qualitative comparison can be found in the supplementary material~\ref{sup:qualitative_results}.

\begin{table}[]
    \centering
    \caption{
    L1, L2, FID, and Flip Ratio (FR) scores for counterfactuals generated by different versions of the proposed approach, compared with results from Jeanneret et al.~\cite{jeanneret2022diffusion}.
    Subscripts \textit{ft} describe LDM instances fine-tuned on ISIC, while \textit{wo-ft} refers to pre-trained LDMs.
    All reported results are computed on the consolidated ISIC test dataset for \textit{MEL} and \textit{NV} classes. 
    The best values are highlighted in bold.
    }    
    \begin{tabular}{l@{\hspace{1.5em}}c@{\hspace{1.5em}}c@{\hspace{1.5em}} c @{\hspace{1.5em}}c}   
        \toprule
        \textbf{Method} & \textbf{L1}  &\textbf{L2} &\textbf{FID}&\textbf{FR}\\ 
        \midrule
        \textbf{DiME}~\cite{jeanneret2022diffusion} & 0.02969  & 0.00176 & 60.79377  &   0.86710        \\
        \textbf{Unconditional \frameworkAcronym$_{wo-ft}$} & 0.01930  &  0.00078 & 17.36892 &  \textbf{1.0}\\
        \textbf{Unconditional \frameworkAcronym$_{ft}$} &  0.01858 & \textbf{0.00073} & 13.45541   & \textbf{1.0}   \\ 
        \textbf{Conditional \frameworkAcronym$_{wo-ft}$} & 0.01930 & 0.00078 & 17.31825 & \textbf{1.0} \\
        \textbf{Conditional \frameworkAcronym$_{ft}$} & \textbf{0.01857}  &\textbf{0.00073}  & \textbf{13.41800}  &\textbf{1.0}    \\ 
        \bottomrule
    \end{tabular}
    \label{tab:fid}
\end{table}

In every version, \frameworkAcronym\ shows a significant improvement in overall metrics, indicating the superiority of counterfactuals generated by LDMs, over traditional approaches.
The fine-tuning of the diffusion model yielded a notable improvement in FID, L1, and L2 measures.
This aligns with the discovery, that the pre-trained LDM is completely incapable of synthesizing skin lesion images (see supplementary material~\ref{sup:image-synthesis}).
The conditional generation of counterfactuals using the target class yielded further minor improvements.
It is notable that in comparison to DiME, \frameworkAcronym\ achieves a perfect flip ratio, indicating a stronger generative capacity of the LDM.
Due to its superior performance, the conditional, fine-tuned \frameworkAcronym\ is chosen for all further experiments. 

\begin{figure}[!h]
    \centering 
    \includegraphics[width=1\linewidth]{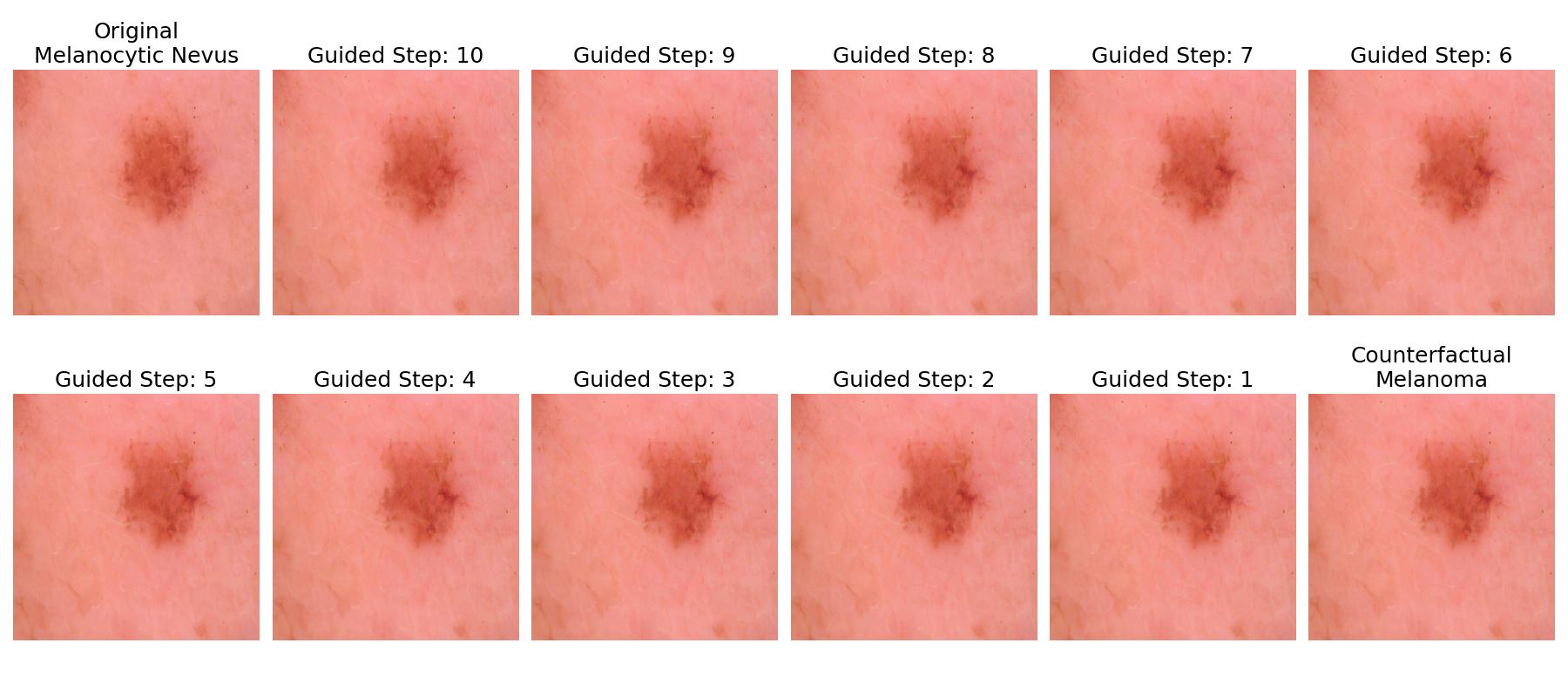}
    \caption{Counterfactual trajectory for a \textit{Nevus} with target class \textit{Melanoma}. Along the process, the manifestation of darker, atypical pigment structures can be observed.}
    \label{fig:counterfactual_trajectories}
\end{figure}

Counterfactual trajectories capture the progression from a factual image to a counterfactual, as shown in figure~\ref{fig:counterfactual_trajectories}. 
The counterfactual trajectory dataset is generated by computing trajectories for each ISIC training image into each alternate target class.
Each trajectory consists of intermediate images from the 10 guided steps as well as the final counterfactual for 7 target classes, resulting in 78 images per sample.
With 23,868 samples in the ISIC training dataset, the resulting counterfactual trajectory dataset contains 1,861,704 samples.

\subsection{Semantic Space Disentanglement}
A VAE was trained on the counterfactual trajectory dataset to capture all variations reflected by the data. 
The encoder comprises six ResNet-based down-sampling blocks and final convolution layers, and the decoder architecture mirrors the encoder executing the reverse process of the encoding phase.
An extensive hyperparameter search has been conducted, using different loss combinations, weightings, batch sizes, learning rates, and corresponding schedules.
The experiments indicate that there is an inherent trade-off between the disentanglement provided by the KLD loss and the reconstruction quality of the VAE.
All reported results are based on a VAE trained for 23 epochs with the Adam optimizer using a batch size of 128, a learning rate of $2.5e^{-5}$, and an exponential decay learning rate scheduler with a decay factor $\gamma$ of 0.95. ISIC test images along with their reconstructions are shown in supplementary material \ref{sup:VAE_results}.

\subsection{Discovery of Relevant Concepts}
\begin{figure*}[htp]
    \centering
    \subfloat[Success Rate = 83.85 \% - Darkening lesion and surrounding skin.]{\label{fig:discovered_dimensions:darkening}
    \includegraphics[width=.95\textwidth]{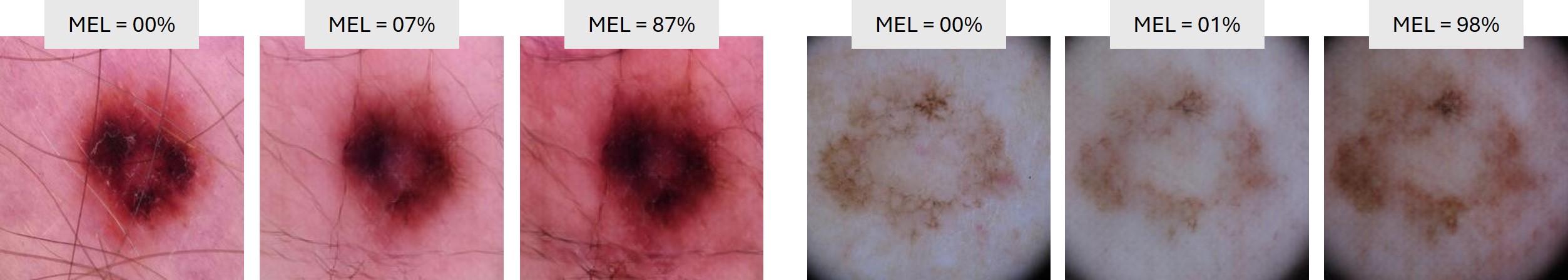}} \\
    \subfloat[Success Rate = 79.99 \% - Darkening lesion and brightening surrounding skin.]{\label{fig:discovered_dimensions:darkinnerwhiteouter}
      \includegraphics[width=.95\textwidth]{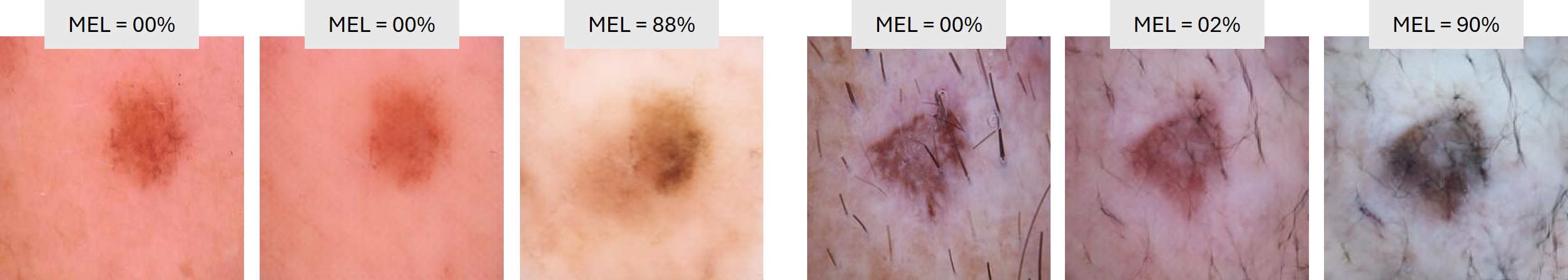}} \\ 
    \subfloat[Success Rate = 72.96 \% - Adding dark corner artifacts.]{\label{fig:discovered_dimensions:dca}
      \includegraphics[width=.95\textwidth]{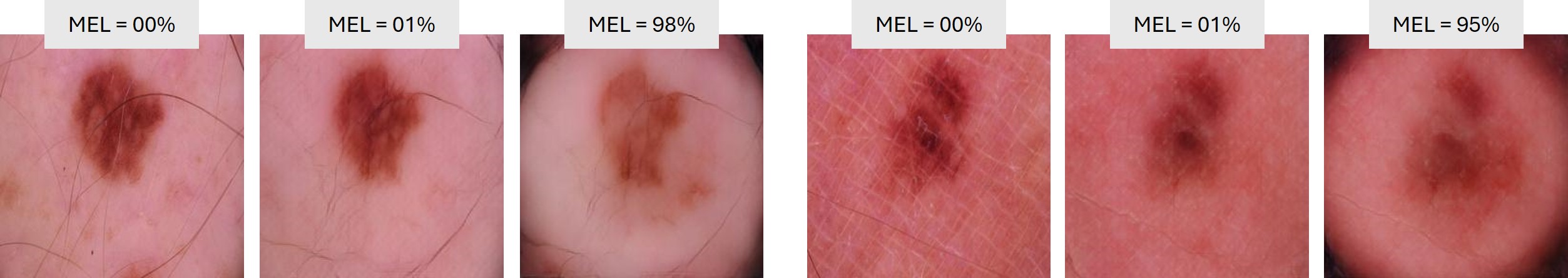}} \\
    \subfloat[Success Rate = 83.74 \% - Reddening of lesion and surrounding skin area.]{\label{fig:discovered_dimensions:rednevus}
      \includegraphics[width=.95\textwidth]{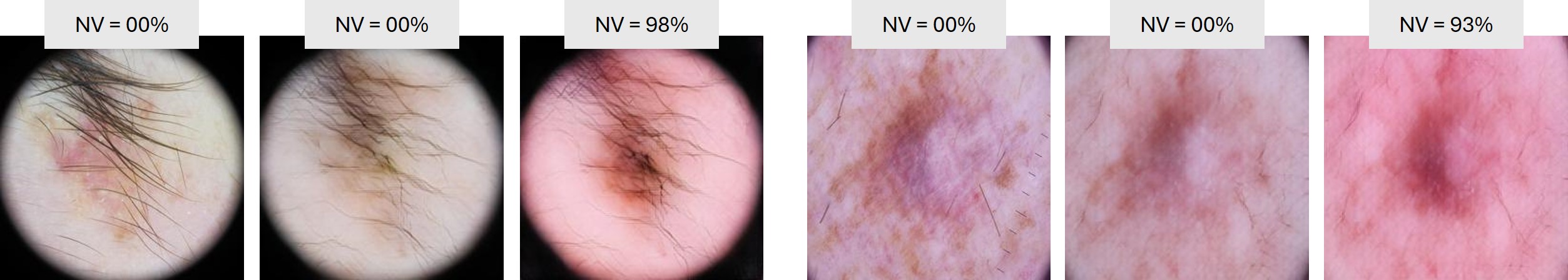}} \\
    \subfloat[Success Rate = 80.27 \% - Whitening of lesion and surrounding skin area.]{\label{fig:discovered_dimensions:whiteningbcc}
      \includegraphics[width=.95\textwidth]{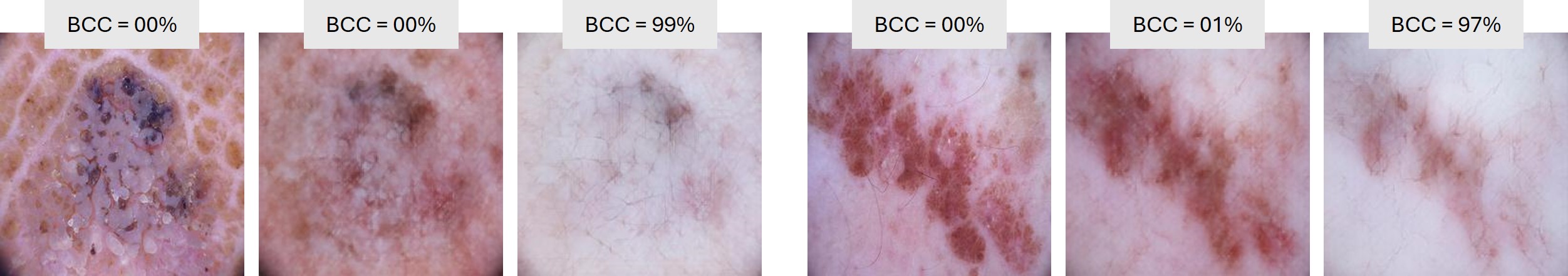}} \\
    \subfloat[Success Rate = 65.18 \% - Adding central white patch on the lesion.]{\label{fig:discovered_dimensions:whitepatchdf}
      \includegraphics[width=.95\textwidth]{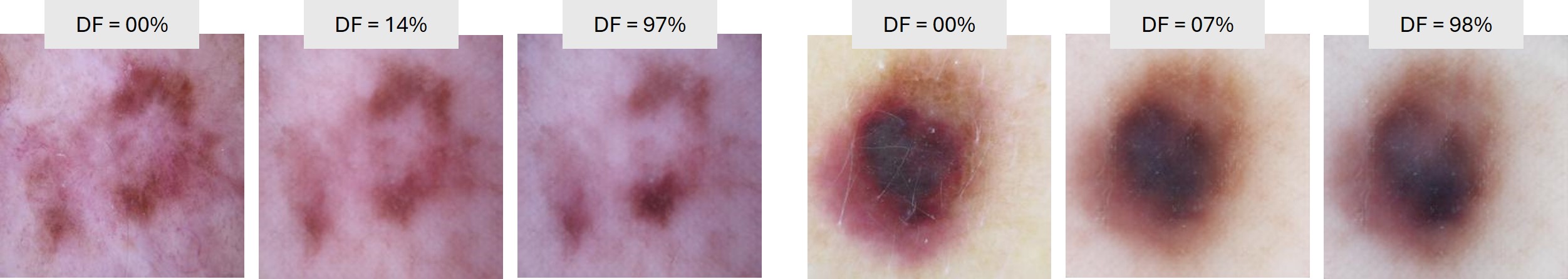}}
    \caption{Examples of discovered concepts using \frameworkAcronym\ with the ISIC dataset. Each row shows two examples of a concept, where the original image, the reconstruction, and the manipulated reconstruction are aligned from left to right.}
    \label{fig:discovered_dimensions}
\end{figure*}

VAE latent dimensions relevant for the target classifier are determined by algorithm~\ref{alg:part3} using the ISIC validation dataset.
Figure~\ref{fig:discovered_dimensions} shows a small selection of the most relevant dimensions identified by \frameworkAcronym.
The success rate of a dimension is provided in the sub-caption for each concept and describes the fraction of test cases where the target class probability increased when altering said dimension.

For \textit{Melanoma}, it has been found that many dimensions are related to darkening the skin lesion area while brightening the surrounding skin (e.g., figure~\ref{fig:discovered_dimensions:darkinnerwhiteouter}).
Moreover, some dimensions (e.g., figure~\ref{fig:discovered_dimensions:darkening}) added darker spots and texture within the skin lesion to promote the prediction of \textit{Melanoma}.
Another dimension adds dark corner artifacts to the images, evident in figure~\ref{fig:discovered_dimensions:dca}.
The inverse of that dimension was found to promote the prediction of \textit{Nevus}.

Similarly, for other classes, mostly concepts related to overall color are identified.
Several relevant dimensions for \textit{Nevus} altered the overall hue of the image to a red color (see figure~\ref{fig:discovered_dimensions:rednevus}).
For \textit{Basal Cell Carcinoma}, the most relevant dimensions were related to the brightening of the whole image (see figure~\ref{fig:discovered_dimensions:whiteningbcc}).
For \textit{Dermatofibroma}, the most relevant dimension added white structures in the center of the lesions (see figure~\ref{fig:discovered_dimensions:whitepatchdf}).
The relevance of the concepts is also emphasized by the high success rates of concepts, as shown in the sub-captions.

\section{Discussion} \label{sec:discussion}

In the previous section, the proposed three-step framework has been successfully applied for the global explanation of a skin lesion classifier.
The results strongly suggest that \frameworkAcronym\ successfully outperforms DiME~\cite{jeanneret2022diffusion} regarding the quality of generated counterfactuals.
Using an LDM allowed for a reduction of the required denoising steps leading to a significant performance improvement in the generation of counterfactuals as compared to DiME (at least 12x speed-up).
Moreover, the reduced number of denoising steps together with the generation in the latent space allowed the omission of the perceptual and the L1 loss employed in \cite{jeanneret2022diffusion} while yielding better counterfactuals.
The conditioning of the LDM for the generation of counterfactual trajectories turned out to yield insignificant improvements in the quality of counterfactuals, but might aid the generation of more diverse and realistic counterfactuals.

Through \frameworkAcronym\, several disentangled dimensions were identified.
The success rates of the presented dimensions confirm their relevance for the target classifier, and therefore their utility for explanation.
By manual inspection, the plausibility of some high-performing dimensions could be established.
For instance, in the case of \textit{Melanoma}, the discovered concept of dark corner artifacts is a bias inherent to ISIC, previously highlighted in the literature as well~\cite{sies2021dark}.
Other dimensions potentially associated with dataset biases include redness, which tends to increase the prediction confidence of \textit{Nevus}, image brightening, which enhances the likelihood of predicting \textit{Basal Cell Carcinoma}, and image darkening, which promotes the probability of \textit{Melanoma} prediction.
Upon nearer investigation of the dataset, it can be observed that these phenomena also reflect in the dataset's statistics (see supplementary material~\ref{sup:colorbias}).

Apart from biases, \frameworkAcronym\ has also yielded concepts that can be clinically proven.
One dimension that led to a particularly high increase in \textit{Dermatofibroma} prediction probability resulted in the introduction of white structures to the center of the lesion.
This concept has been already documented in medical literature as a white, scar-like structure~\cite{agero2006conventional} that is highly indicative of the diagnosis, but not yet proven in DL-based classifiers. 

\section{Limitations \& Future Work}
\label{sec:limitations}
Although \frameworkAcronym\ was successfully used to discover previously unknown concepts in this work, some limitations remain.
Despite the automated selection of classifier-relevant dimensions, a persisting challenge in concept discovery is the manual interpretation of concepts that often requires the help of domain experts.
Future works might address this limitation by providing more sophisticated analysis and visualization methods that segment, aggregate, and cluster changes induced by the manipulation of concept dimensions.
Moreover, the exploration of new concepts is drastically hampered by the existence of significant biases in the dataset.
The linear process of concept discovery might have to be replaced in practice by an iterative approach that involves a debiasing step, which would facilitate the discovery of meaningful concepts and improve the classifier's validity.
Finally, the utilization of the VAE architecture for deriving a disentangled representation of counterfactual trajectories might be further improved in the future.
The VAE training presented a crucial trade-off between reconstruction fidelity and disentanglement through the weight of the KLD loss.
As skin lesion classification relies on fine-grained features like subtle textures, reconstruction was favored over disentanglement.
This, however, led to one entangled dimension that stood out for its comparatively high relevance throughout all classes, introducing inconclusive changes to the images.
Future work should address this issue by applying more sophisticated disentanglement techniques with better fine-grained reconstruction capability, such as StyleGAN.

\section{Conclusion}
\label{sec:conclusion}

In high-stakes scenarios such as healthcare, DL models play a pivotal role in early disease detection, potentially saving lives.
It is essential to provide human-aligned explanations which offer insights into the complex decision-making of models to gain the trust of their users.
These insights can potentially reveal new biomarkers relevant in clinical practice.
This study provides a new automated framework for the discovery of concepts based on the synthesis of counterfactual trajectories using LDMs.

The proposed \frameworkAcronym\ framework is the first to use LDMs with classifier guidance to generate counterfactual explanations.
Its proposed counterfactual generation step yields better FID scores compared to the previous DiME~\cite{jeanneret2022diffusion} method while being up to 12 times more resource-efficient.
A counterfactual trajectory dataset is constructed, reflecting relevant semantic changes along the decision boundaries of the target classifier.
A disentangled representation of these classifier-relevant cues is derived using a VAE.
The automatic and unsupervised exploration of this latent representation yielded valuable insights into the decision-making behavior of a skin lesion classifier, revealing not only biases but also previously unknown biomarkers, supported by first evidence in the medical literature.
\frameworkAcronym\ can be applied to arbitrary application domains such as radiology and histology, providing a combination of local and global explanations, and therefore paving the way to trustworthy AI finding its way into clinical practice. 

\section{Acknowledgements}
The project was funded by the Federal Ministry for Education and Research (BMBF) with grant number 03ZU1202JA.

\bibliographystyle{splncs04}
\bibliography{references}
\clearpage

\appendix
\section{Supplementary Material}

\subsection{Color Biases in the ISIC dataset}
\label{sup:colorbias}
\begin{figure}[H]
    \centering
    \includegraphics[width=1\linewidth]{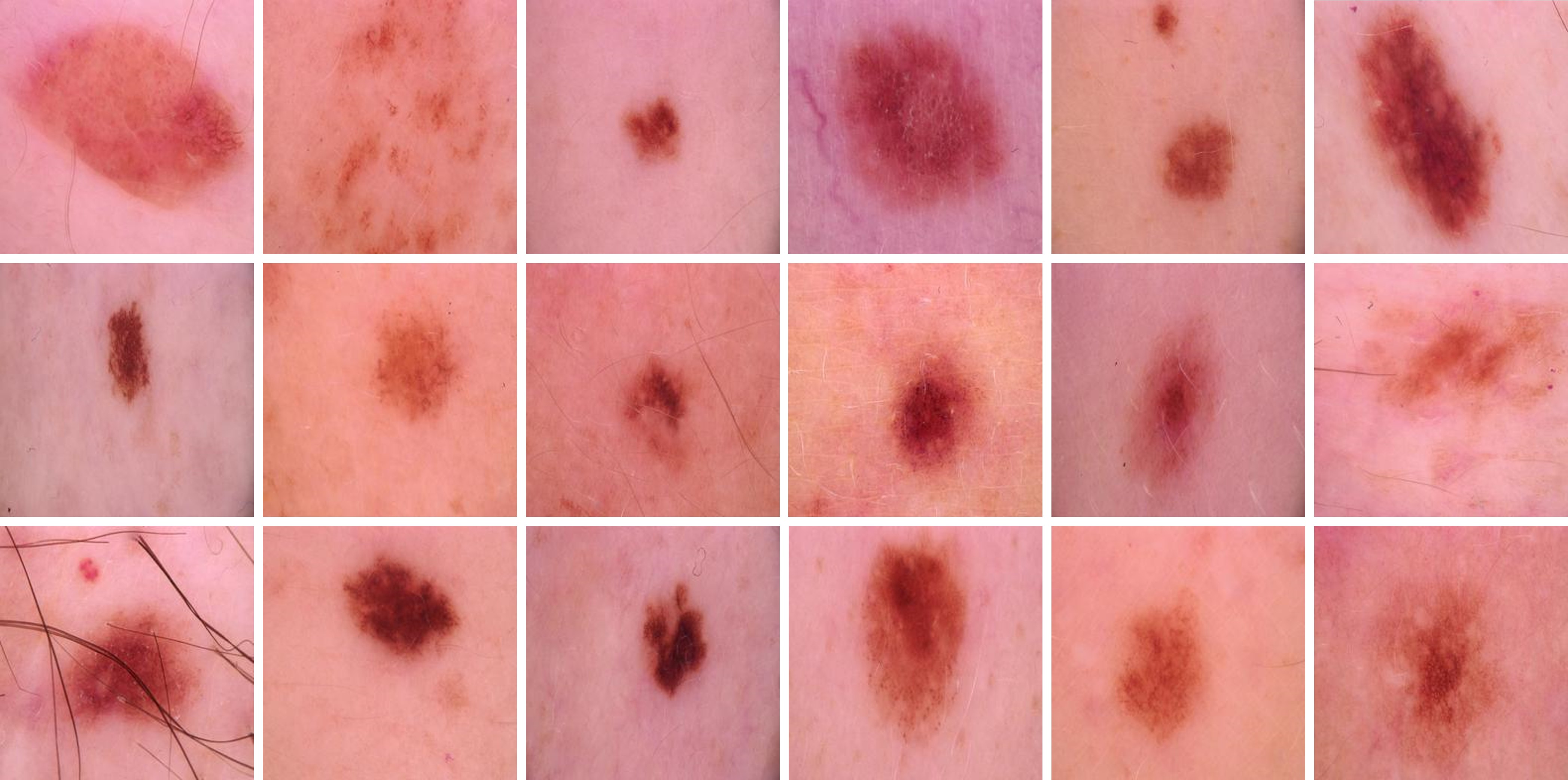}
    \caption{Exemplary nevus images from ISIC with typical red hue.}
    \label{fig:app:rednevus}
\end{figure}

The concept discovery using the proposed \frameworkAcronym\ framework highlighted that color plays a pivotal role in the classification of ISIC.
This suggests the presence of biases in the dataset.
Table~\ref{tab:ISICColorBiases} shows the average values of red (in RGB) and saturation (in HSV) channels of images from the different ISIC classes.
The values have been computed separately for the lesion region as well as the surrounding skin using a segmentation algorithm.
It can be observed, that the \textit{Basal Cell Carcinoma} has the lowest average saturation value on both lesion and skin and that the skin in ISIC's \textit{Nevus} images exhibit the highest average value for the channel red.
This provides first strong evidence for the existence of color-related biases in ISIC.
Exploratory analysis of the dataset further reveals that there are large subsets of the \textit{Nevus} class, that exhibit a similar, reddish color hue (see figure~\ref{fig:app:rednevus}).

\begin{table}[]
    \centering
    \caption{Average color values for red (in RGB space) and saturation (in HSV space) for the ISIC dataset.}
    \label{tab:ISICColorBiases}
    \begin{tabular}{@{}clccccccc@{}}
        \toprule
        \multicolumn{1}{l}{\textbf{Area}}    &  \textbf{Value}   & \textbf{NV}            & \textbf{DF}            & \textbf{VASC}          & \textbf{MEL}           & \textbf{BCC}           & \textbf{BKL}           & \textbf{AKIEC}         \\
        \midrule
        \multirow{2}{*}{\textbf{Lesion}} & Red & 169 & \textbf{193} & 171 & 152 & 189 & 167 & 194 \\
                                & Saturation & 122 & 081 & 101 & 094 & \textbf{067} & 073 & 075 \\
        \multirow{2}{*}{\textbf{Skin}}   & Red & \textbf{208} & 201 & 205 & 188 & 196 & 181 & 199 \\
                                & Saturation & 070 & 052 & 057 & 046 & \textbf{046} & 048 & 055 \\
        \bottomrule
    \end{tabular}

\end{table}

\clearpage
\subsection{Generation of Synthetic Samples}
\label{sup:image-synthesis}
\begin{figure}[H]
    \centering 
     \subfloat{\includegraphics[width=0.87\textwidth]{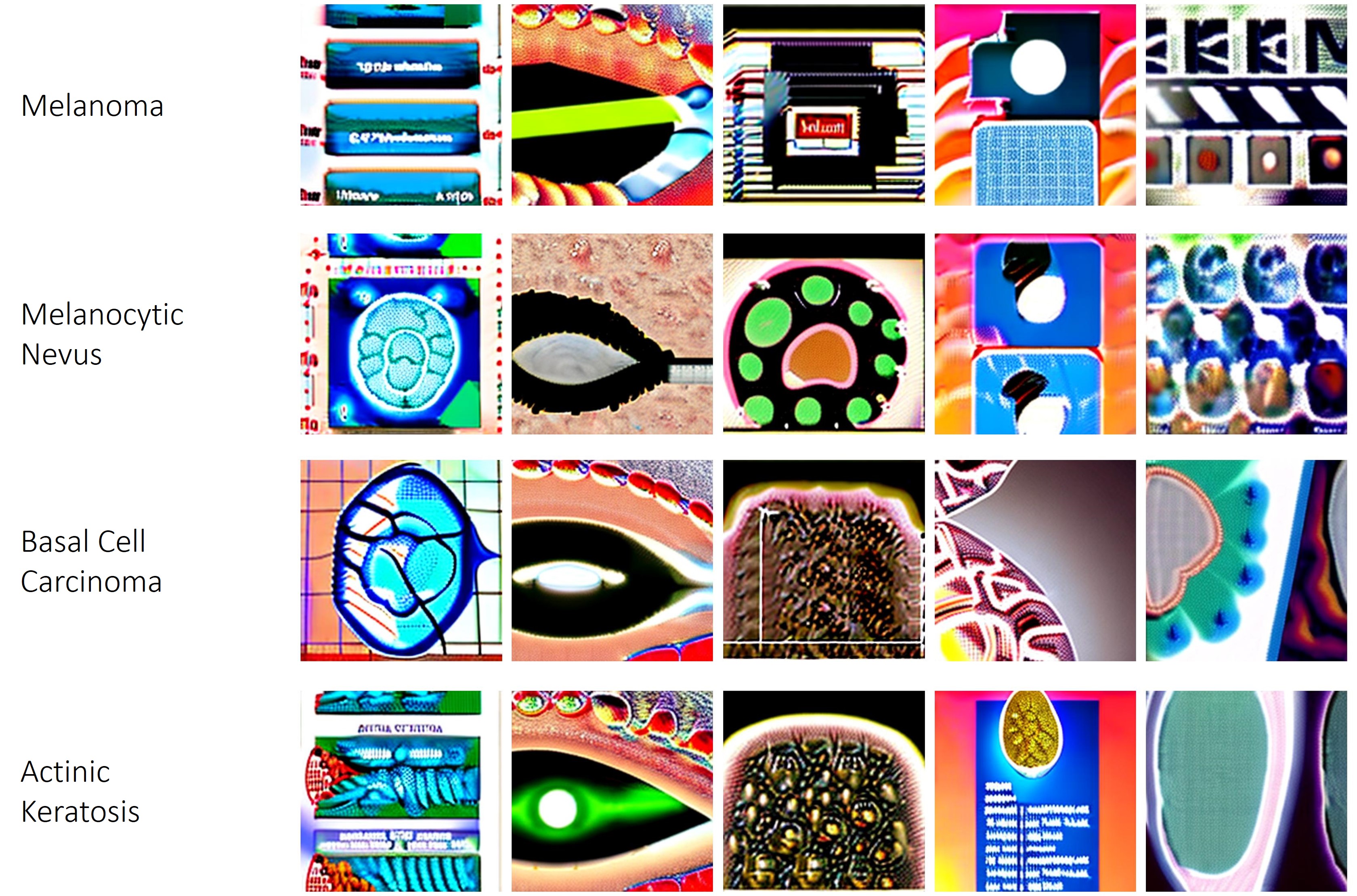}} \\
     \vspace{-0.7 em}
      \subfloat{\includegraphics[width=0.87\textwidth]{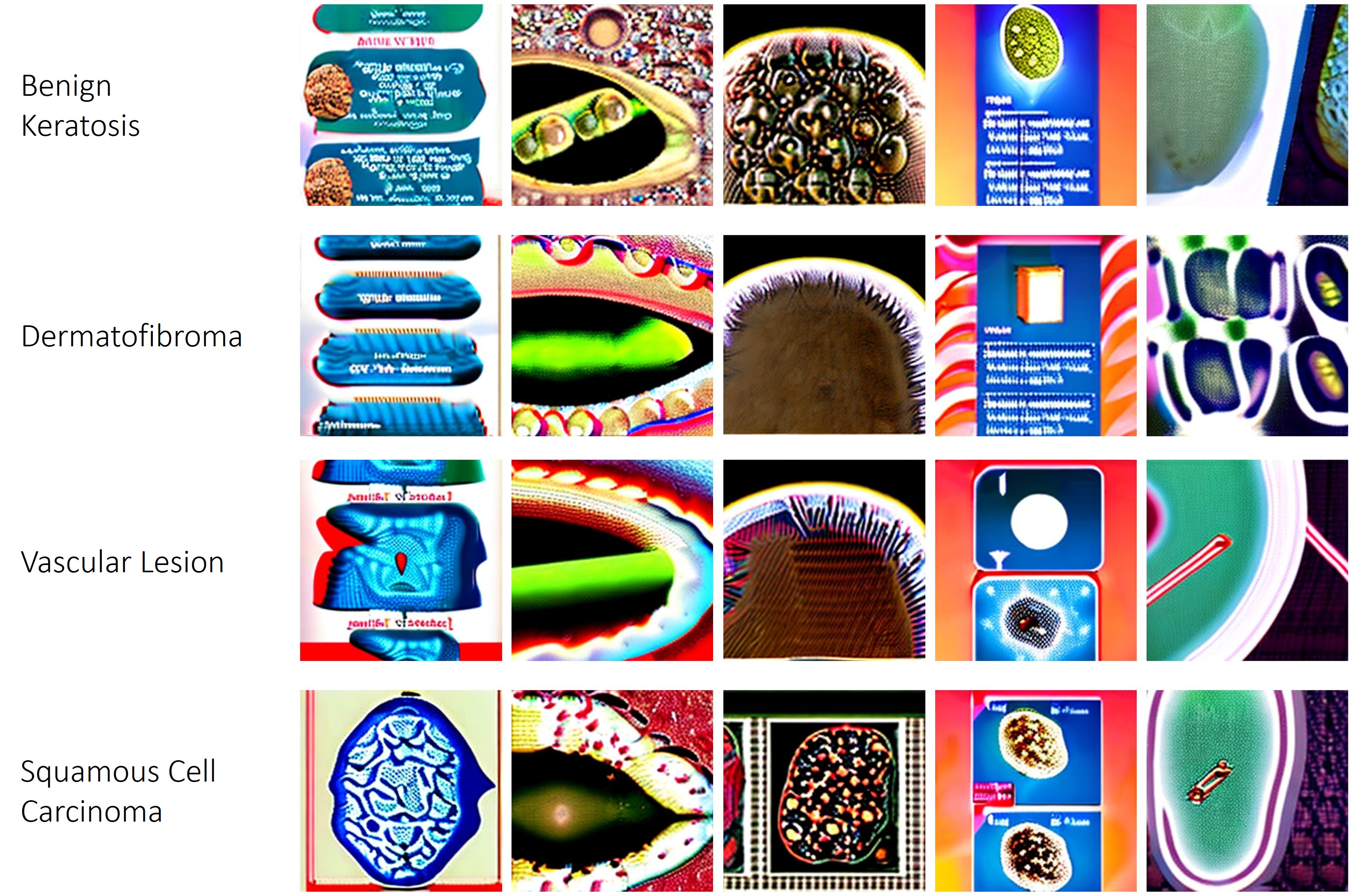}} \\
      \caption{Synthetic images produced by a pre-trained LDM. The text prompt utilized for generating samples for each skin class label is shown on the left side.} 
     \label{fig:sd_sample_wo_ft} 
\end{figure}

Synthetic samples corresponding to skin lesion classes were generated utilizing a pre-trained LDM. 
The generation process involved using the class names as prompts to produce the samples.
However, it was observed that the LDM, without being fine-tuned on a skin lesion dataset, cannot generate realistic images of skin lesions as shown in figure~\ref{fig:sd_sample_wo_ft}.

\begin{figure}[H]
    \centering 
     \subfloat{\includegraphics[width=0.87\textwidth]{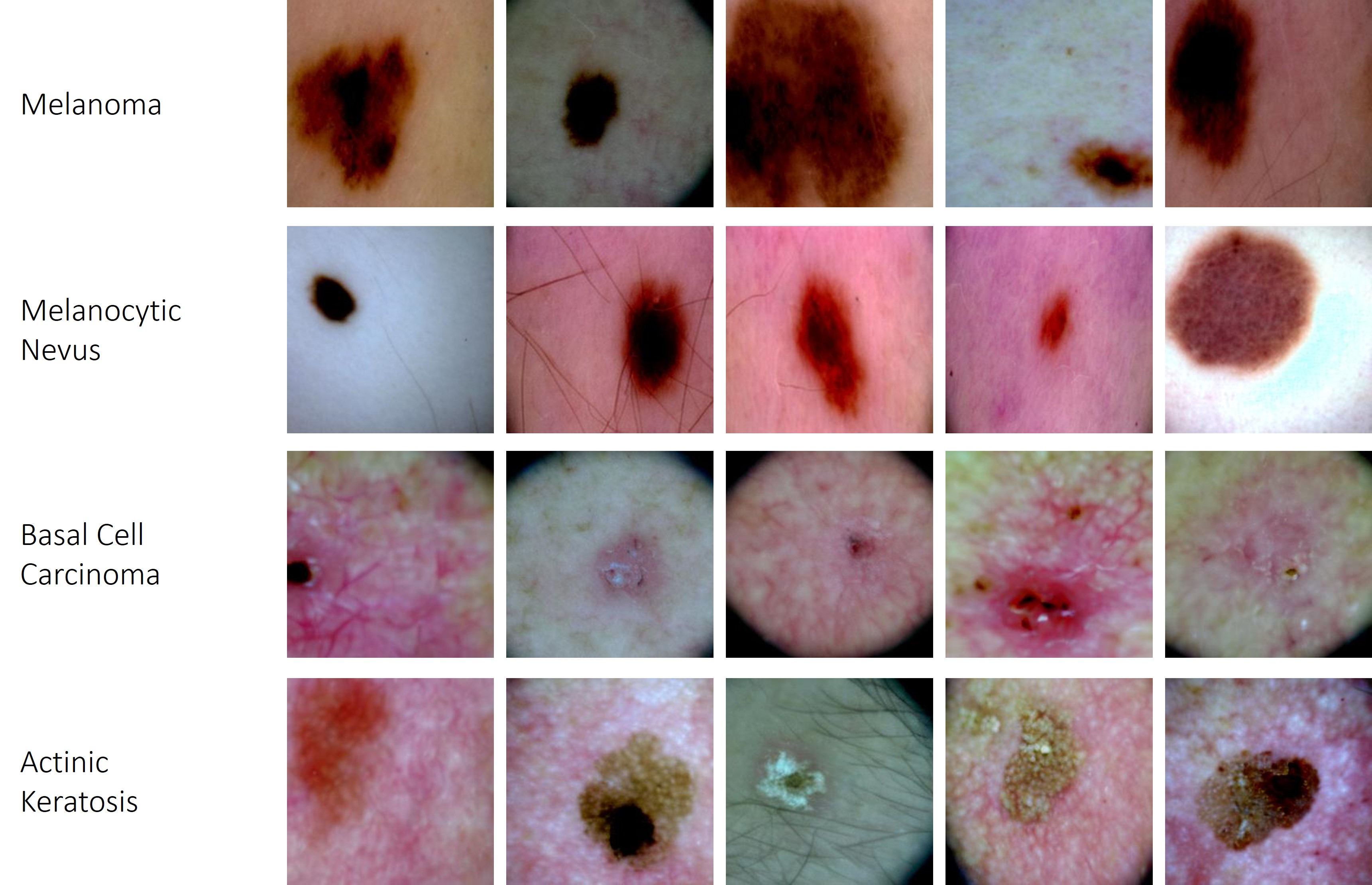}} \\
     \vspace{-0.7 em}
      \subfloat{\includegraphics[width=0.87\textwidth]{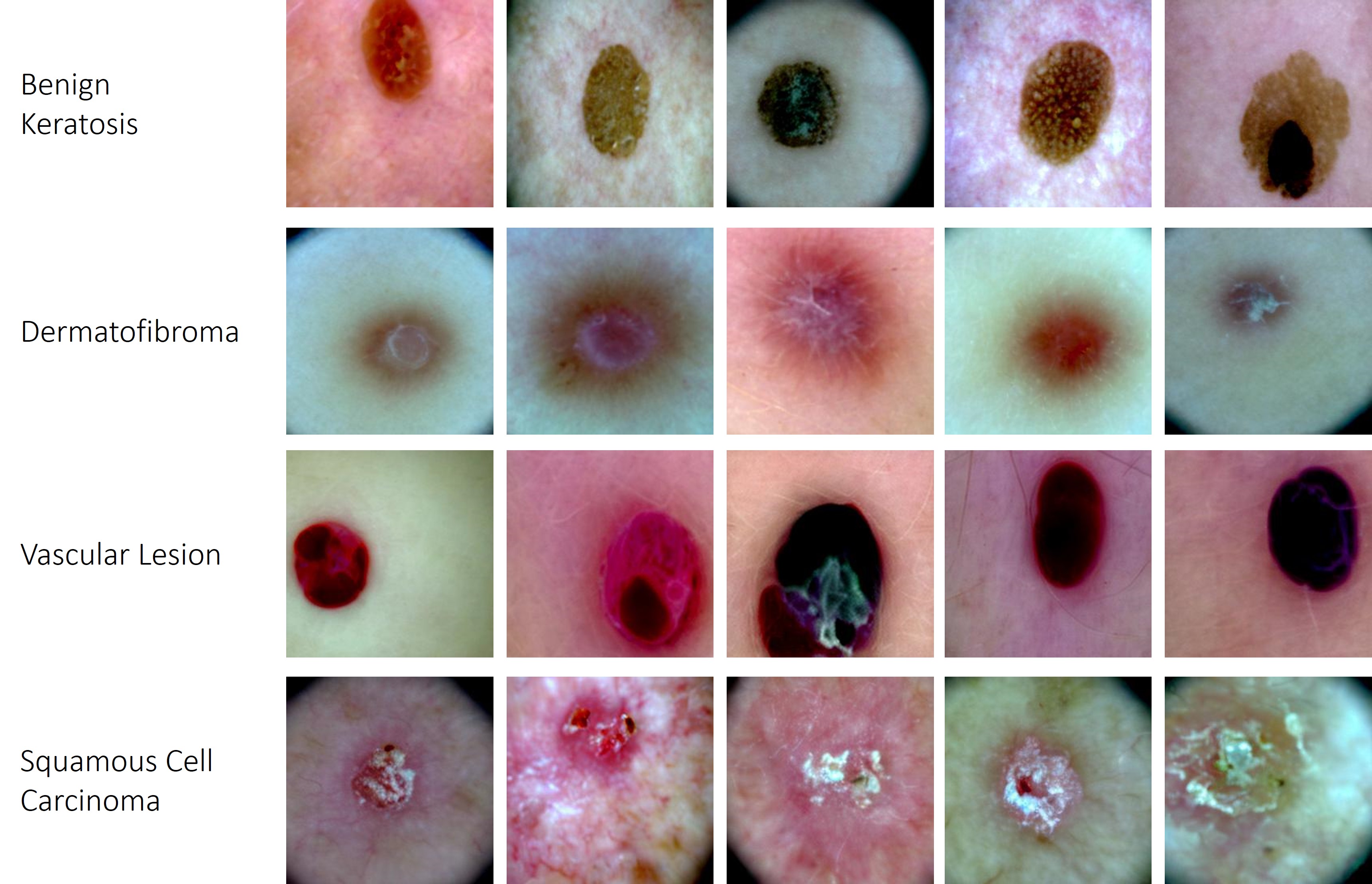}} \\
      \caption{Synthetic images produced by the LDM fine-tuned on the consolidated ISIC challenge dataset. The text prompt utilized for generating samples for each skin class label is shown on the left side.} 
     \label{fig:sd_sample} 
\end{figure}

To address this limitation, the LDM was subsequently fine-tuned using the consolidated training dataset from the ISIC challenge. After fine-tuning, samples were generated using text prompts corresponding to the specific classes of skin lesions, with the resultant images being shown in figure~\ref{fig:sd_sample}.

\clearpage

\subsection{Counterfactual Generation Results}
\label{sup:counterfactuals}
\begin{figure}[h!]
    \centering
    \includegraphics[width=1\linewidth]{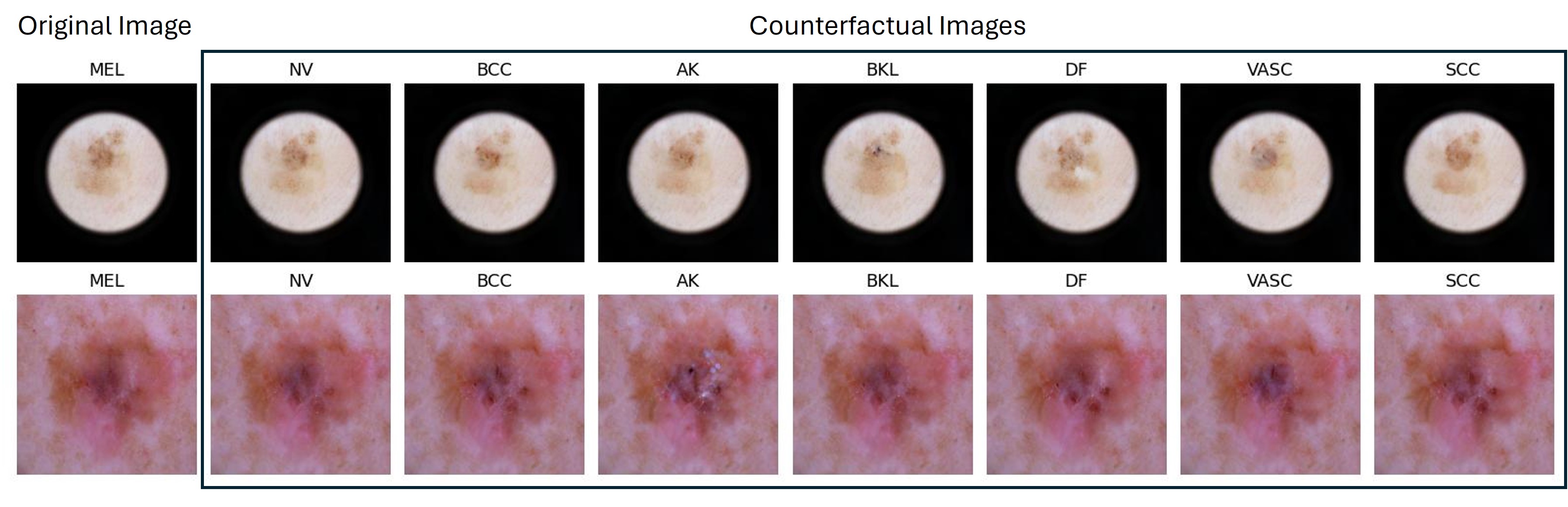}
    \caption{\textbf{Melanoma} to all other target classes.}
    \label{fig:MEL_all}
\end{figure}

\begin{figure}[h!]
    \centering
    \includegraphics[width=1\linewidth]{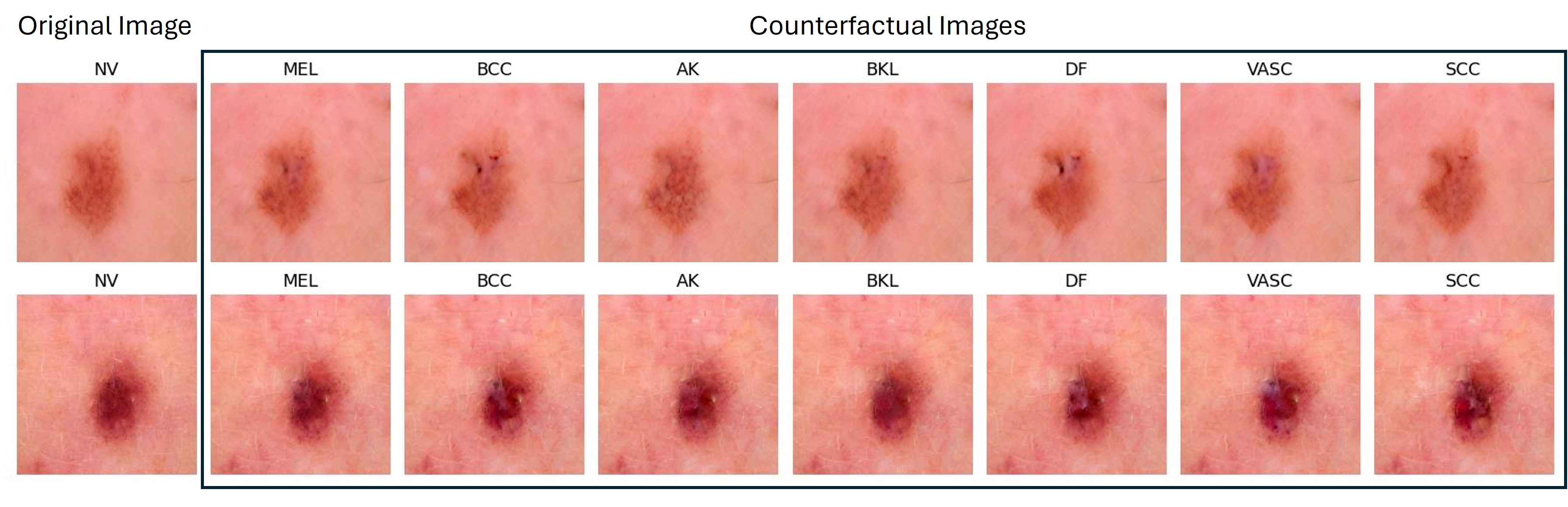}
    \caption{\textbf{Nevus} to all other target classes.}
    \label{fig:NV_all}
\end{figure}

\begin{figure}[h!]
    \centering
    \includegraphics[width=1\linewidth]{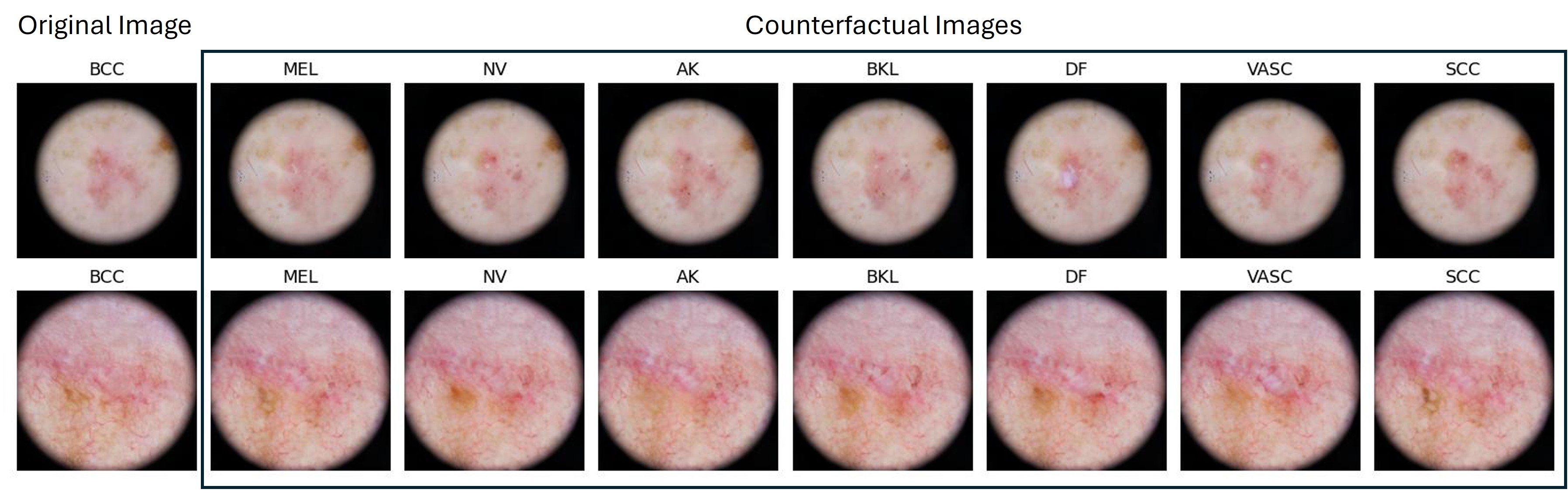}
    \caption{\textbf{Basal Cell Carcinoma} to all other target classes.}
    \label{fig:BCC_all}
\end{figure}

\begin{figure}[h!]
    \centering
    \includegraphics[width=1\linewidth]{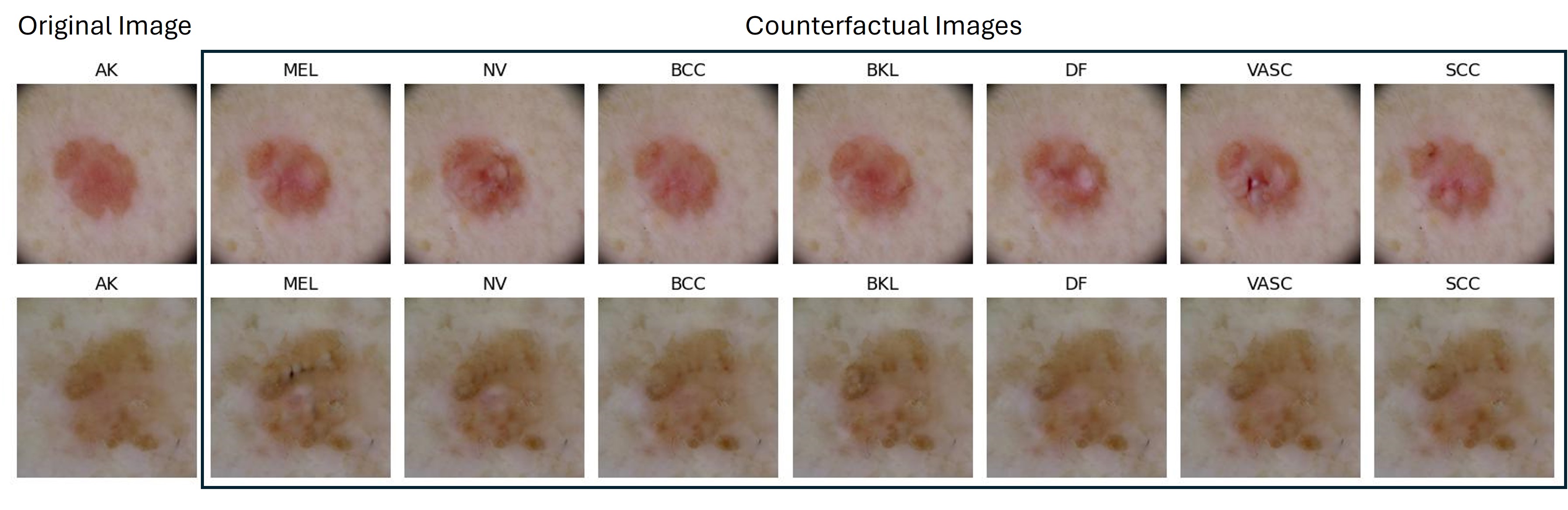}
    \caption{\textbf{Actinic Keratosis} to all other target classes.}
    \label{fig:AK_all}
\end{figure}

\begin{figure}[h!]
    \centering
    \includegraphics[width=1\linewidth]{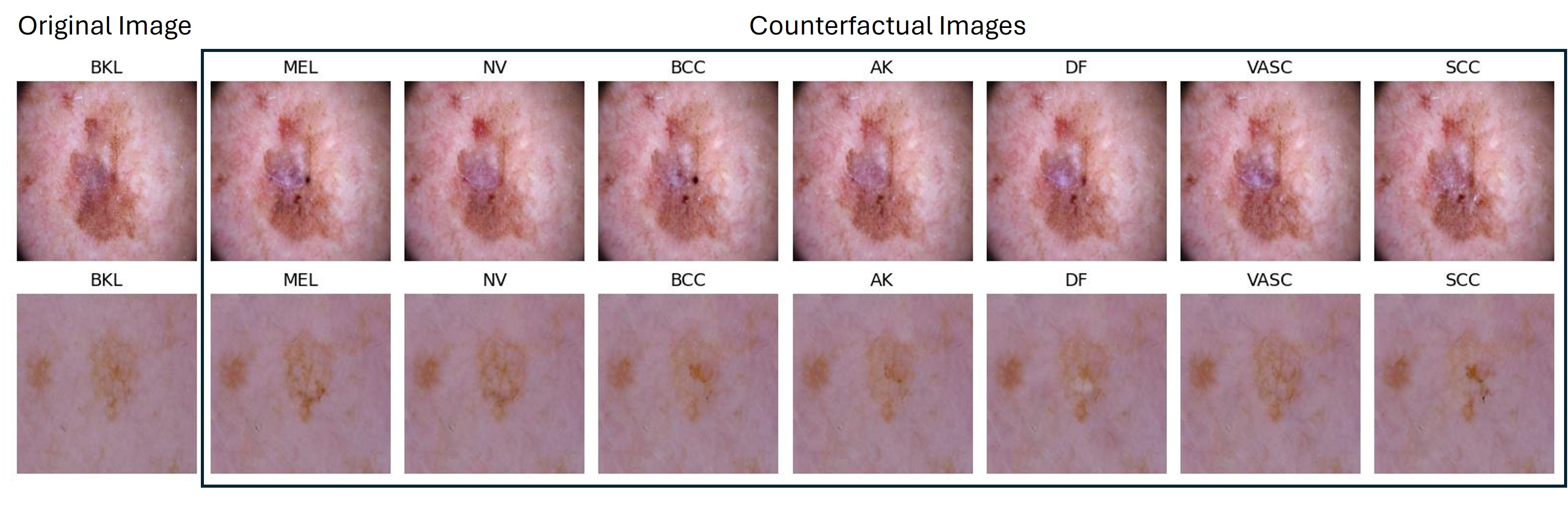}
    \caption{\textbf{Benign Keratosis} to all other target classes.}
    \label{fig:BKL_all}
\end{figure}

\begin{figure}[h!]
    \centering
    \includegraphics[width=1\linewidth]{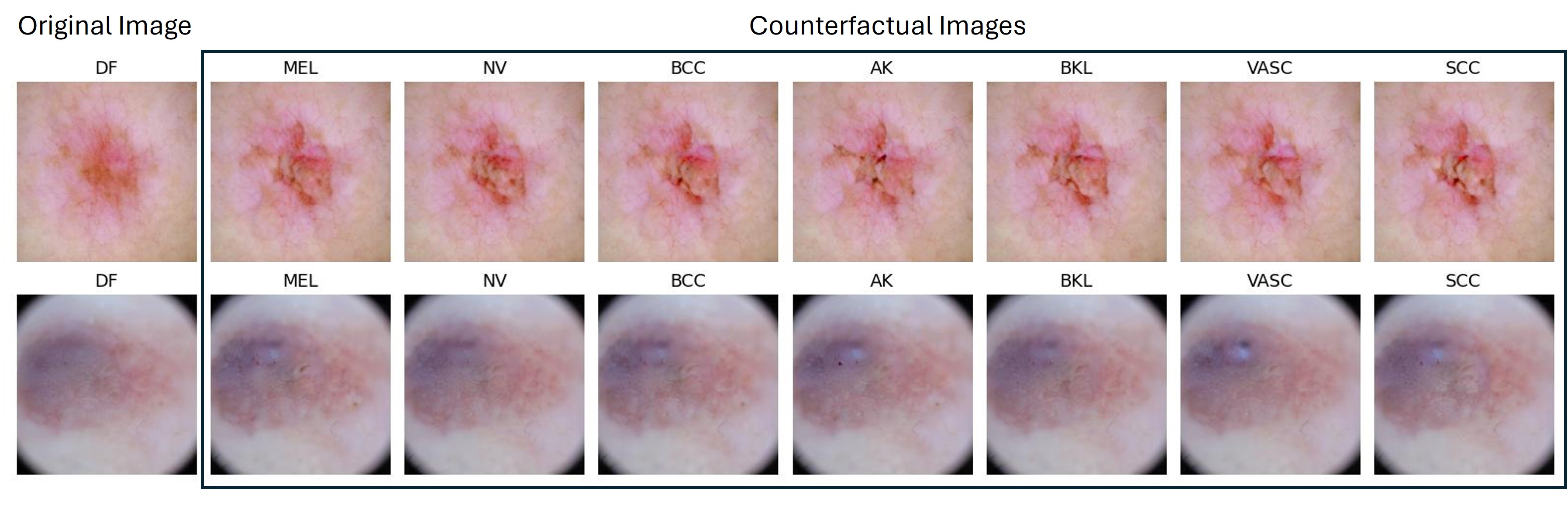}
    \caption{\textbf{Dermatofibroma} to all other target classes.}
    \label{fig:DF_all}
\end{figure}

\begin{figure}[h!]
    \centering
    \includegraphics[width=1\linewidth]{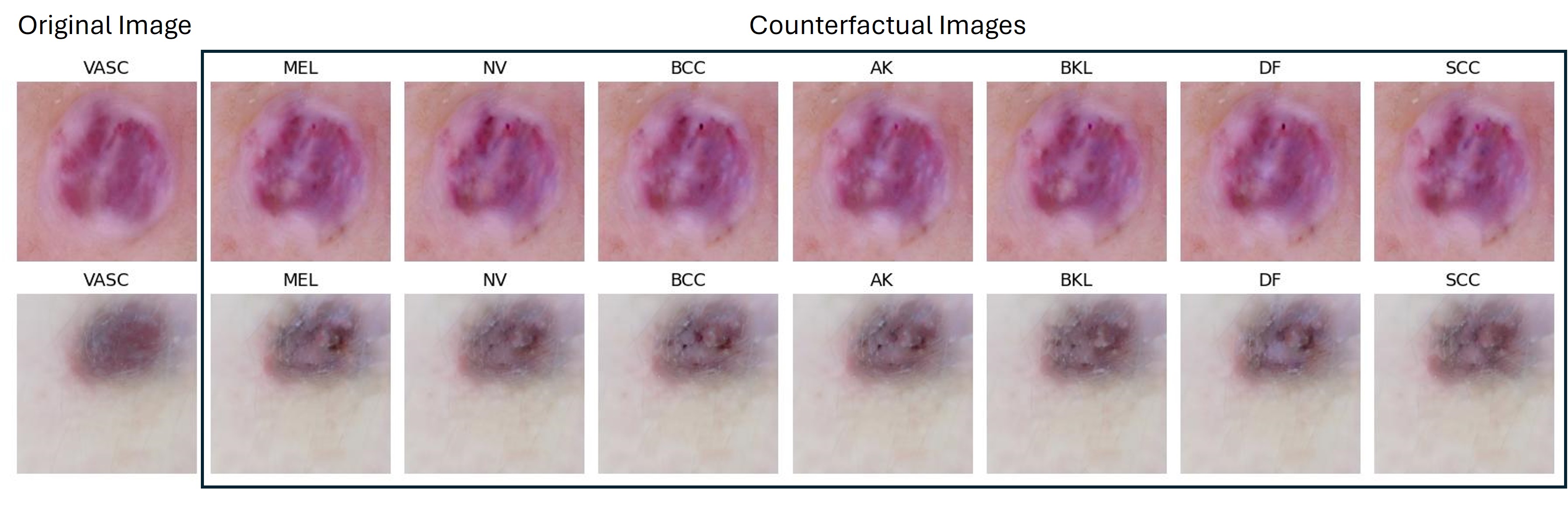}
    \caption{\textbf{Vascular Lesion} to all other target classes.}
    \label{fig:VASC_all}
\end{figure}

\begin{figure}[h!]
    \centering
    \includegraphics[width=1\linewidth]{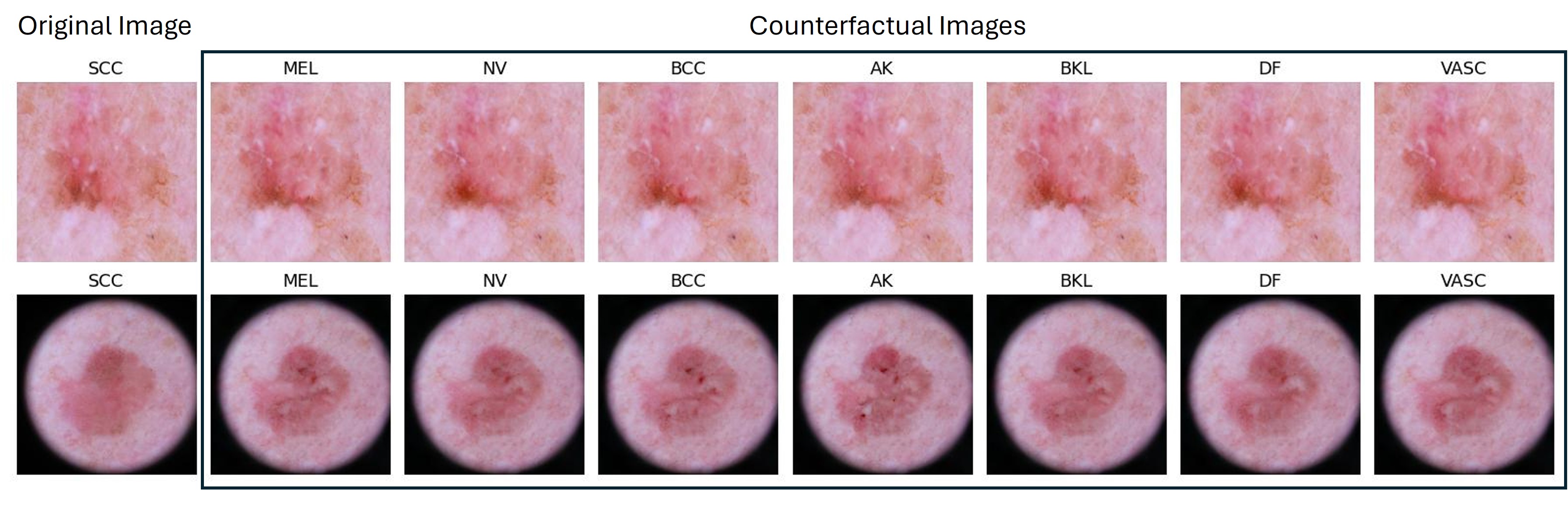}
    \caption{\textbf{Squamous Cell Carcinoma} to all other target classes.}
    \label{fig:SCC_all}
\end{figure}

Counterfactuals can be created from a single source class to each target class, showcasing the distinct variations in the generated images for each class. To comprehensively illustrate these differences, we generate counterfactuals from one class to all other classes. 
For instance, if a sample image is initially classified as \textit{Melanoma}, counterfactuals for this image can be generated for each of the remaining target classes, which include \textit{Melanocytic Nevus}, \textit{Basal Cell Carcinoma}, \textit{Actinic Keratosis}, \textit{Benign Keratosis}, \textit{Dermatofibroma}, \textit{Vascular Lesion}, and \textit{Squamous Cell Carcinoma}, effectively illustrating the discriminative features of each class.
Figures \ref{fig:MEL_all} to \ref{fig:SCC_all} display the counterfactual images generated from each source class showing the distinctive variations when transitioning across all target classes. 

\clearpage
\subsection{Comparison of Qualitative Result}
\label{sup:qualitative_results}
\begin{figure}[H]
    \centering
    \includegraphics[width=1\linewidth]{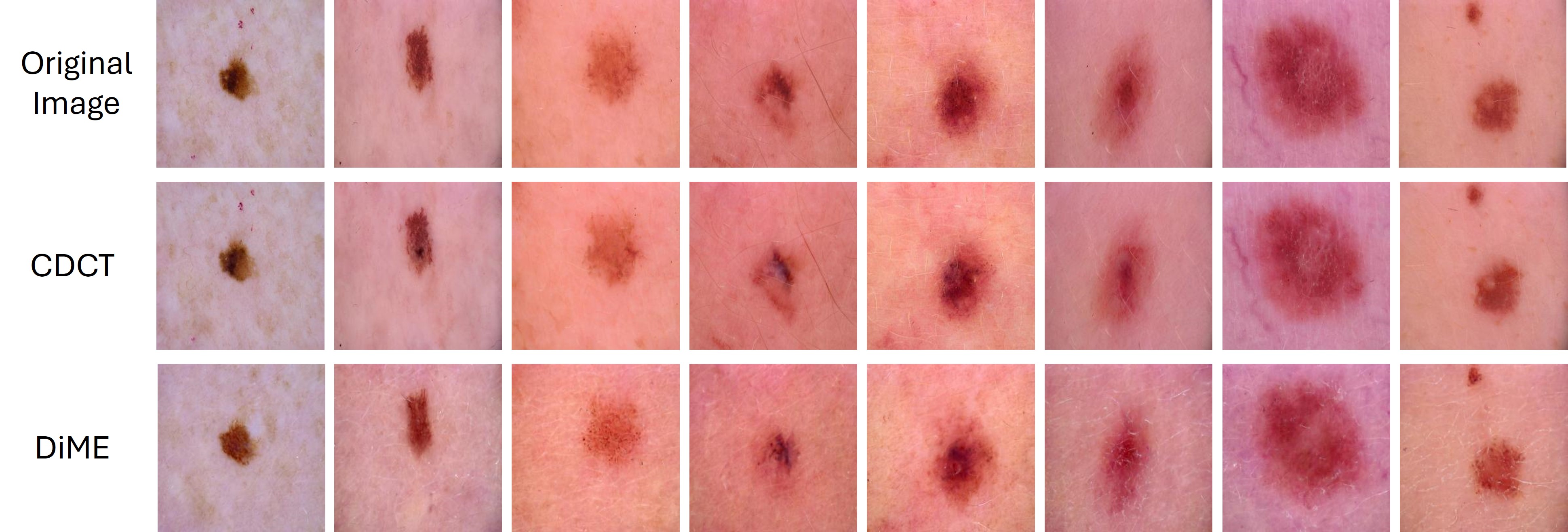}
    \caption{Comparison of counterfactuals generated for eight exemplary images from the \textit{Nevus} class to \textit{Melanoma} using \frameworkAcronym\ and DiME~\cite{jeanneret2022diffusion}.}
    \label{fig:sup:qualitative_nev-to-mel}
\end{figure}

\begin{figure}[H]
    \centering
    \includegraphics[width=1\linewidth]{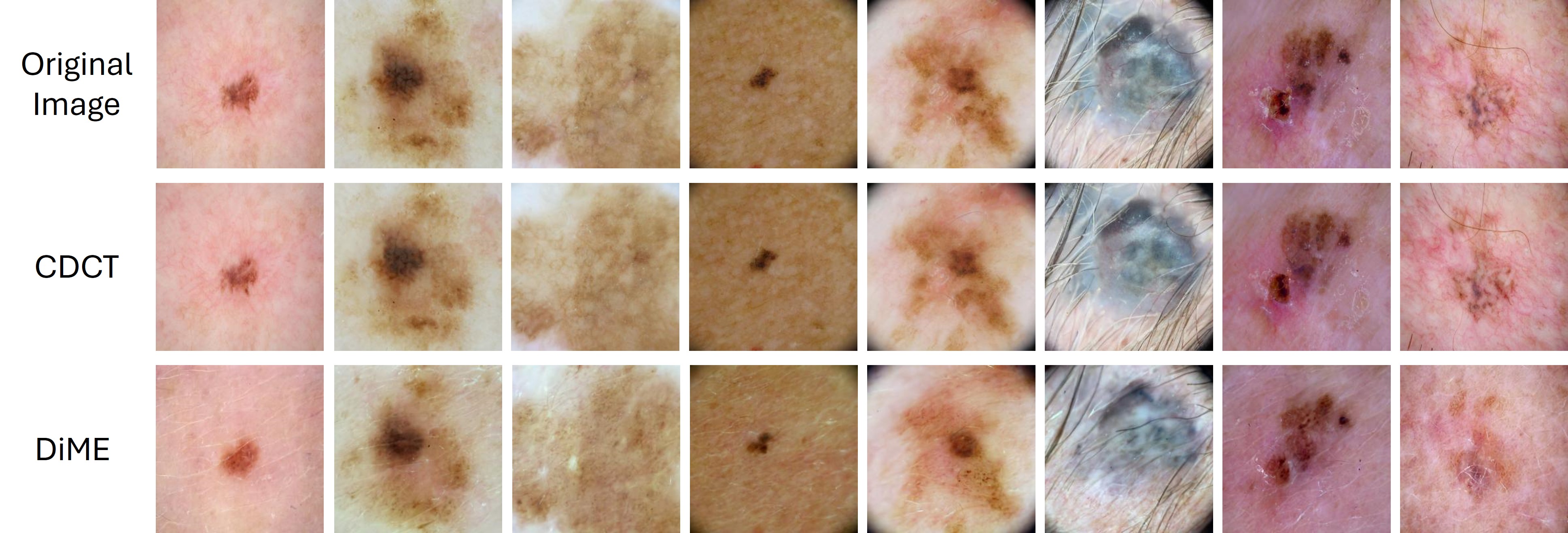}
    \caption{Comparison of counterfactuals generated for eight exemplary images from the \textit{Melanoma} class to \textit{Nevus} using \frameworkAcronym\ and DiME~\cite{jeanneret2022diffusion}.}
    \label{fig:sup:qualitative_mel-to-nev}
\end{figure}

Figure~\ref{fig:sup:qualitative_nev-to-mel} and \ref{fig:sup:qualitative_mel-to-nev} show a qualitative comparison of counterfactual results generated with \frameworkAcronym\ and  the previous state-of-the-art method DiME~\cite{jeanneret2022diffusion}.

\clearpage
\subsection{VAE reconstruction Results}
\begin{figure}
    \centering
    \includegraphics[width=0.9\linewidth]{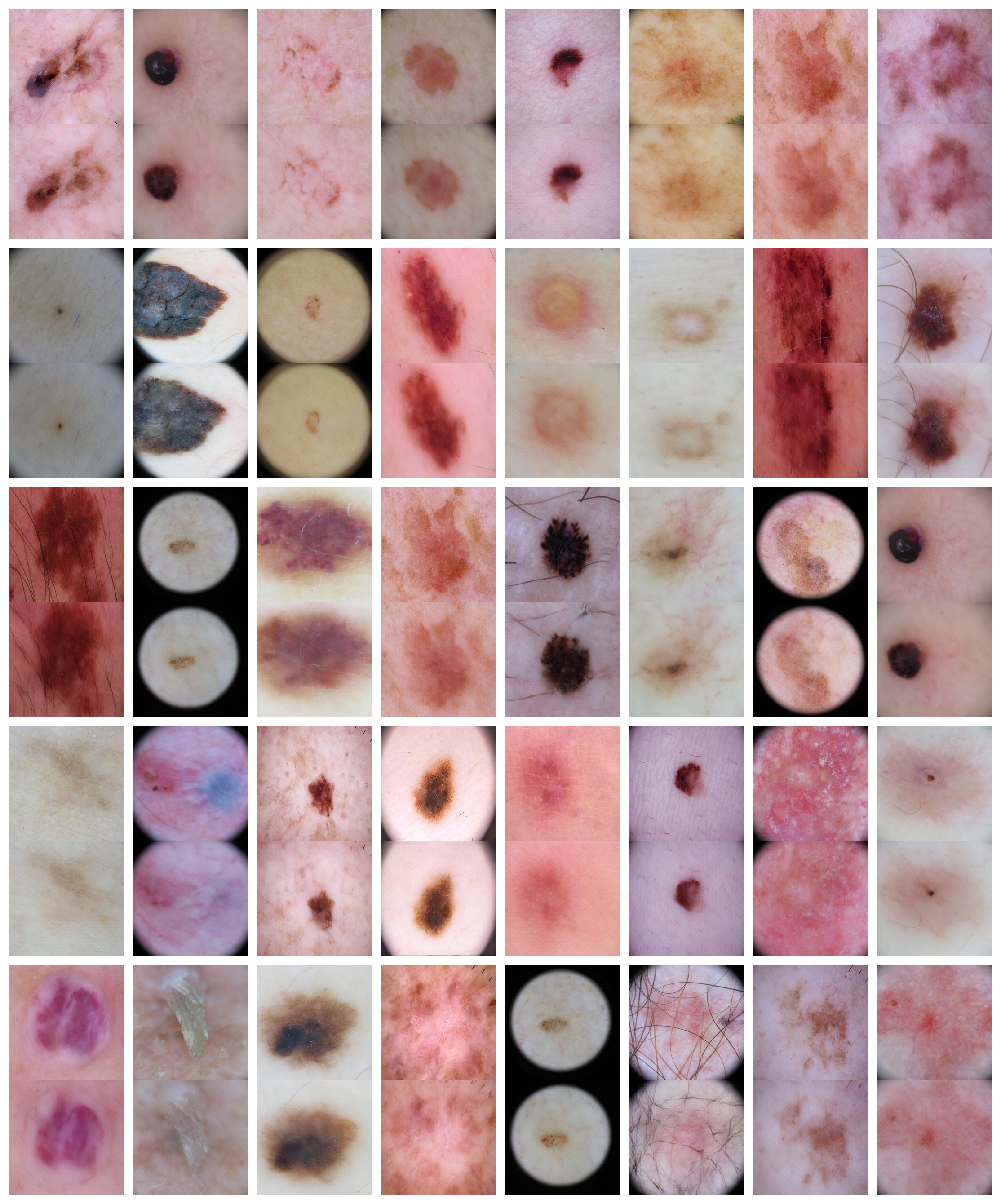}
        \caption{Pairs of random ISIC test images (top) and their corresponding reconstructed image (bottom) generated by the trained VAE.}
    \label{fig:vae_reconstruction}
\end{figure}

Figure \ref{fig:vae_reconstruction} depicts the reconstructed images generated by the trained VAE. While the reconstruction captures a substantial amount of information, it is noteworthy that the process does not entirely preserve detailed information. While conveying the essence of the original data, the reconstructed images exhibit a level of abstraction where finer details may not be reproduced. Despite losing some fine-grained details, the reconstructed images demonstrate the VAE's ability to reproduce critical characteristics from the training data.
\label{sup:VAE_results}

\end{document}